\documentclass[10pt,twocolumn,letterpaper]{article}

\usepackage{xspace}
\usepackage{booktabs}
\usepackage{multirow}
\usepackage{colortbl}
\usepackage{xcolor}
\usepackage{times}
\usepackage{epsfig}
\usepackage{graphicx}
\usepackage{amsmath}
\usepackage{amssymb}
\usepackage{hyperref}
\usepackage[margin=1in]{geometry} 
\usepackage[numbers,sort&compress]{natbib}
\usepackage{caption}

\newcommand{\modelname}{BBQ\xspace}

\begin{document}

\title{BBQ-to-Image: Numeric Bounding Box and Qolor Control in Large-Scale Text-to-Image Models}
\author{
Eliran Kachlon \ \ \ Alexander Visheratin \ \ \ Nimrod Sarid \ \ \ Tal Hacham \ \ \  Eyal Gutflaish \\ Saar Huberman \ \ \ Hezi Zisman \ \ \ David Ruppin \ \ \  Ron Mokady \ \ \ 
\\[2mm]
\vspace{1em}
BRIA AI 
\\[0.5mm]
\vspace{-40pt}
\centering
}
\date{} 

\twocolumn[{
\vspace{-30pt}
\maketitle
\begin{center}
\begin{minipage}{\textwidth}
\centering
\includegraphics[width=\textwidth]{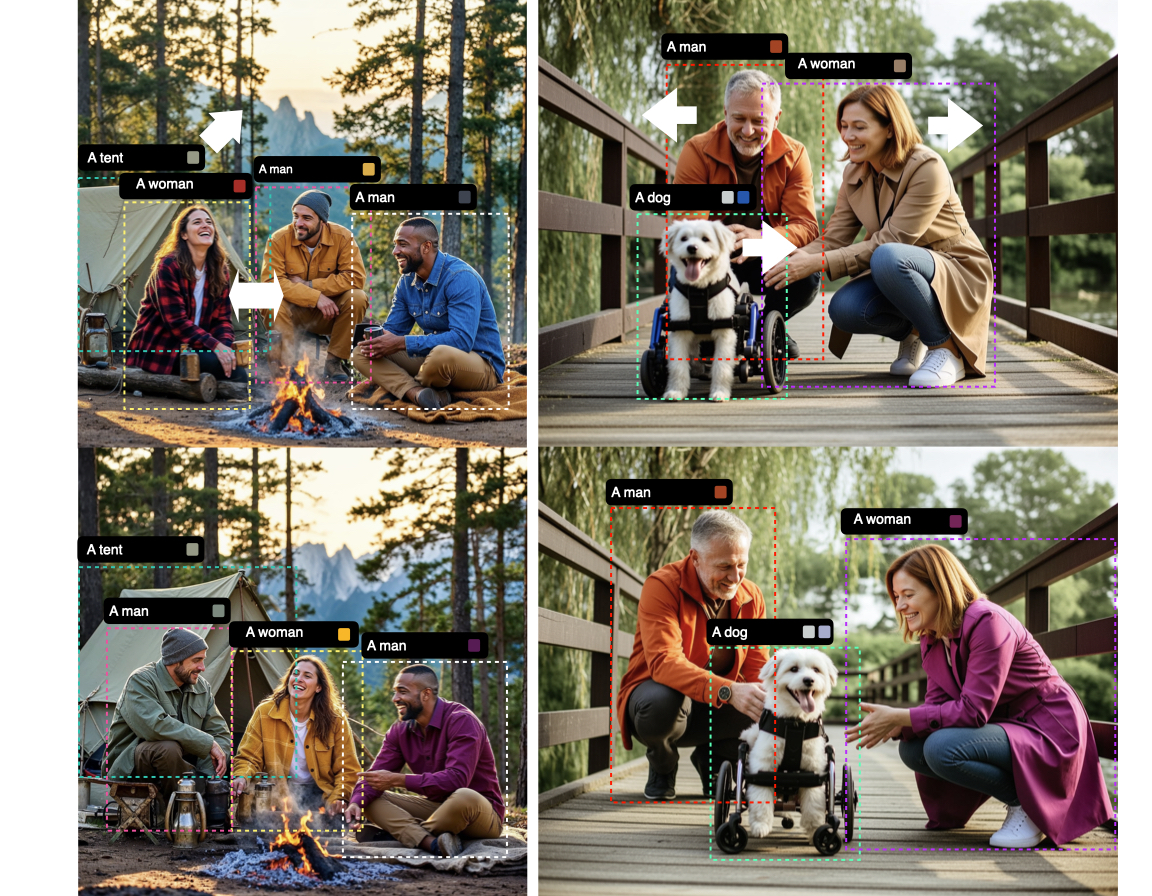}
\captionof{figure}{\textbf{Bounding-box and RGB-controlled image generation and refinement.}
BBQ enables precise spatial and color control by conditioning on explicit numeric bounding boxes and RGB values. In the example, the exact locations of the people and the dog are specified via bounding boxes, and the colors of their clothing are defined using RGB triplets. 
Beyond initial generation, BBQ enables structured refinement by modifying only the numeric parameters in the caption and re-generating the image. Due to the model’s disentangled control over layout and color, updating bounding boxes (e.g., swapping the man and the woman, or moving the dog to the right) or modifying RGB values results in consistent, targeted changes while preserving the rest of the scene.}

\label{fig:teaser}
\end{minipage}
\end{center}
\vspace{40pt}
}]


\begin{abstract}
Text-to-image models have rapidly advanced in realism and controllability, with recent approaches leveraging long, detailed captions to support fine-grained generation. 
However, a fundamental \emph{parametric gap} remains: existing models rely on descriptive language, whereas professional workflows require precise numeric control over object location, size, and color.
In this work, we introduce \emph{BBQ}, a large-scale text-to-image model that directly conditions on numeric bounding boxes and RGB triplets within a unified structured-text framework. 
We obtain precise spatial and chromatic control by training on captions enriched with parametric annotations, without architectural modifications or inference-time optimization. 
This also enables intuitive user interfaces such as object dragging and color pickers, replacing ambiguous iterative prompting with precise, familiar controls.
Across comprehensive evaluations, BBQ achieves strong box alignment and improves RGB color fidelity over state-of-the-art baselines. 
More broadly, our results support a new paradigm in which user intent is translated into an intermediate structured language, consumed by a flow-based transformer acting as a renderer and naturally accommodating numeric parameters.
\end{abstract}
\section{Introduction}\label{sec:intro}

\begin{figure*}[t]
  \centering
  \includegraphics[width=\textwidth]{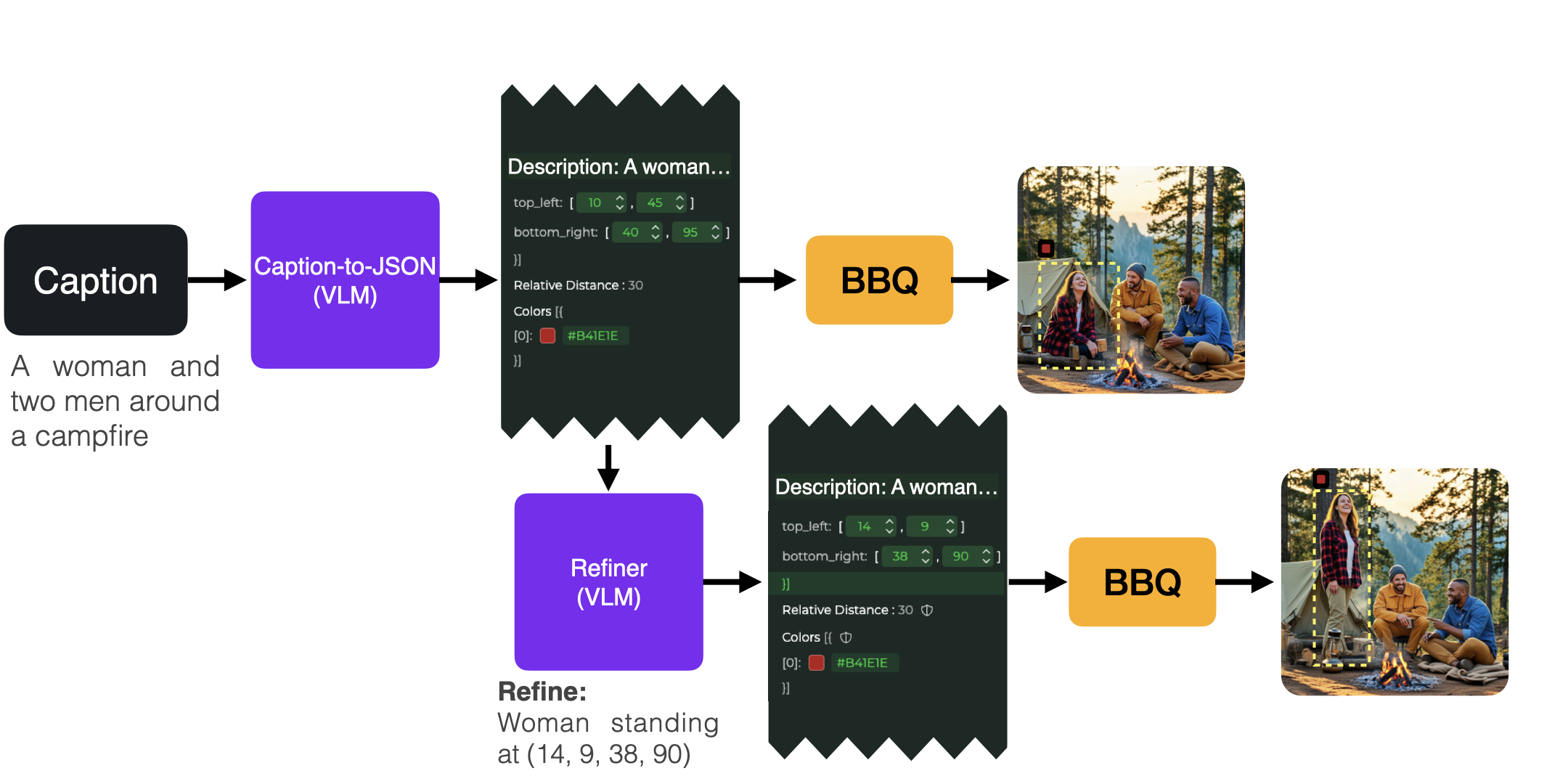}
\caption{\textbf{End-to-end parametric workflow.}
A short prompt is expanded by a VLM into a structured JSON that includes numeric bounding boxes and RGB values (for clarity, we show only the parametric fields for the woman).
The JSON is then provided to \modelname{} to generate an image. Users can edit specific fields (e.g., box coordinates or color values), and \modelname{} updates the output accordingly while preserving unrelated content, demonstrating native disentanglement.
Notably, \modelname{} receives no image input, and consistency is maintained solely through the disentangle structured conditioning.}

  \label{fig:workflow}
\end{figure*}

Text-to-image models have rapidly evolved from casual creative tools into professional-grade systems, achieving unprecedented levels of realism and visual fidelity. 
Recent works have significantly advanced controllability by training on long structured captions, most notably \emph{FIBO}~\cite{gutflaish2025generating}, as well as concurrent systems such as Hunyuan 3.0 \cite{cao2025hunyuanimage} and FLUX.2 \cite{batifol2025flux}. 
By encoding fine-grained visual attributes explicitly in text, these models allow users to specify and control nearly every aspect of an image using language alone. 
Unlike earlier approaches, such models exhibit natural disentanglement, enabling refinement of a specific visual factor, such as lighting, object appearance, or expression, while keeping other aspects unchanged.

Despite this progress, a fundamental \emph{parametric gap} remains. 
Text-based controllability is inherently descriptive and imprecise for attributes that require exact numeric specification. 
In this work, we focus on three such attributes: \emph{size}, \emph{location}, and \emph{color}. 
Current models rely on subjective linguistic descriptors such as ``crimson'' or ``bottom-right,'' whereas professional workflows demand deterministic precision in the form of explicit RGB values and pixel-accurate bounding boxes. 
Moreover, parametric grounding naturally enables intuitive interaction: bounding boxes support direct object manipulation (e.g., dragging), and RGB values integrate seamlessly with color pickers. 
This replaces ambiguous natural-language prompting with precise and familiar user interfaces.

In this paper, we show that large-scale text-to-image models can be adapted to process \emph{numeric inputs} for precise parametric control.
We introduce \emph{\modelname}, a large-scale text-to-image model capable of controlling Bounding Boxes and Qolors directly.
Unlike prior approaches, \modelname{} requires no architectural modifications, no special grounding tokens, and no inference-time optimization.
Instead, parametric control is achieved solely by augmenting the training captions, resulting in a simple yet powerful solution that scales naturally to professional use.

To generate training data, we augment FIBO-style structured captions with explicit numeric attributes, including RGB color values and object bounding boxes.
For inference, we fine-tune a vision–language model (VLM) to serve as an inference-time bridge, converting short natural-language prompts into detailed parametric descriptions that \modelname{} can execute faithfully.

More broadly, our framework highlights a new paradigm for image generation.
Rather than generating images directly from user-written text, user intent is first translated, by a VLM, into an intermediate, structured language, which is then consumed by a flow-based transformer acting as a renderer.
Within this paradigm, we show that the intermediate language can naturally accommodate numeric parameters, enabling precise, deterministic control without sacrificing expressiveness.

Through extensive evaluation, we demonstrate that \modelname{} achieves strong results in precision for object location, size, and color control, demonstrating that large-scale text-to-image models can natively process numeric parameters within a unified text-based framework. 

\section{Related Works}\label{sec:related-works}

\paragraph{Text-to-image models.}
Diffusion models have become the primary framework for text-to-image generation. Early models~\cite{nichol2021glide, saharia2022photorealistic, ramesh2022hierarchical} established the power of conditioning on strong language encoders, while latent diffusion made large-scale training practical~\cite{rombach2022high, podell2023sdxl}. Recently, architectures have shifted toward transformer backbones and flow-matching objectives~\cite{esser2024scaling, flux2024, liu2024playground, cai2025hidream, wu2025qwenimagetechnicalreport}. Together, these advances have pushed the boundaries of visual fidelity.

\paragraph{Long and structured captions.} 
While early models relied on noisy, web-scraped data~\cite{schuhmann2022laion}, recent works~\cite{betker2023improving, esser2024scaling, liu2024playground} show that descriptive synthetic captions significantly improve prompt alignment. 
Recently, FIBO~\cite{gutflaish2025generating} extended this approach by using vision-language models to produce long, structured JSON captions that capture all visual factors in the image, including object attributes, spatial relations, and photographic style. This approach achieved state-of-the-art prompt alignment and introduced fine-grained control, enabling ``native disentanglement'' where modifying a single attribute in the JSON affects only the intended visual factor.
However, FIBO relies on natural language (e.g., ``red'' or ``top-left'') that still involves semantic ambiguity. \modelname builds on this foundation by replacing descriptive strings with absolute precision, integrating RGB values and bounding boxes to transition from semantic alignment to exact pixel-level and chromatic controllability.

\paragraph{Region-controlled text-to-image.}
Traditional layout-to-image frameworks~\cite{zhao2019image, sun2019image, li2020bachgan, frolov2021attrlostgan, li2021image, rombach2022high, yang2022modeling, fan2023frido} that generate images given bounding boxes are usually limited to constrained vocabularies~\cite{lin2014microsoft}.
Recent works like ReCo~\cite{yang2023reco}, GLIGEN~\cite{li2023gligen}, InstanceDiffusion~\cite{wang2024instancediffusion} and Ranni~\cite{feng2024ranni} mitigate these gaps by introducing specialized position tokens or modifying model architectures to inject regional grounding signals. Related controllable diffusion frameworks such as ControlNet~\cite{zhang2023controlnet} and Composer~\cite{huang2023composer} further enable spatial control through additional conditioning pathways. Training-free approaches such as BoxDiff~\cite{xie2023boxdiff} and MultiDiffusion~\cite{bartal2023multidiffusion} also support region constraints by altering the denoising process at inference time. While effective, these approaches necessitate complex structural changes, auxiliary conditioning mechanisms, or inference-time modifications.
In contrast, \modelname unifies high-precision spatial control within a single structured textual representation, enabling exact coordinate guidance without any architectural modifications to the underlying model.

\paragraph{Color-palette generation.}
Controlling color distribution is a classical challenge in image synthesis~\cite{reinhard2002color, welsh2002transferring, levin2004colorization, chang2015palette, aharoni2017pigment}.
Early deep learning efforts incorporated  generative and adversarial frameworks to better model realistic and diverse color distributions~\cite{zhang2016colorful,zhang2017real,lei2019fully,su2020instance, wu2021towards,iizuka2016let,bahng2018coloring,  wang2022palgan}. More recent approaches attempt to provide fine-grained control over color attributes within text-to-image diffusion models. These include methods
that train and fine-tune existing models~\cite{feng2024ranni, butt2024colorpeel, huang2023composer}, as well as training-free 
approaches~\cite{lobashev2025color, shukla2024test, laria2025leveraging} that enable color control by manipulating the sampling process or exploiting existing semantic bindings, bypassing the need for additional fine-tuning. 
However, these methods often rely on specialized adapters, task-specific loss functions, or additional inference-time optimization steps. 
In contrast, \modelname achieves precise RGB-level color attribution by encoding explicit RGB triplets directly within the textual conditioning, without introducing architectural changes or inference-time modifications.

\definecolor{hydrantpink}{RGB}{212,106,140}
\definecolor{catred}{RGB}{255,3,3}
\definecolor{womanred}{RGB}{204,1,1}
\definecolor{shirtblue}{RGB}{0,50,98}

\setlength{\tabcolsep}{0.5pt}
\renewcommand{\arraystretch}{1.0}

\begin{figure}[t]
\centering
\newcommand{\imgw}{0.24\linewidth} 
\begin{tabular}{@{}cccc@{}}
 
  \includegraphics[width=\imgw]{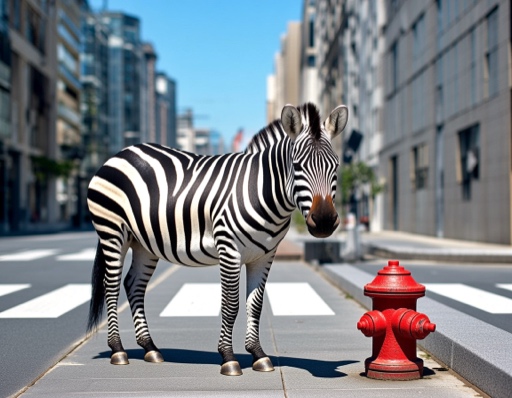} &
  \includegraphics[width=\imgw]{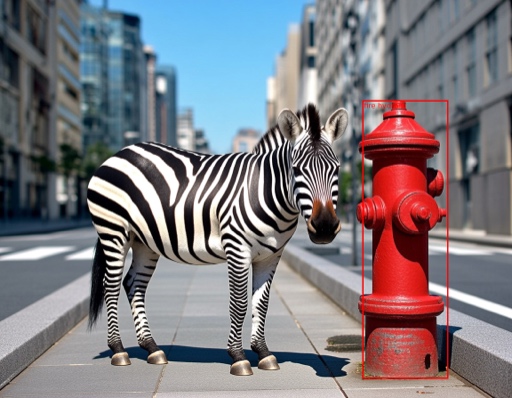} &
  \includegraphics[width=\imgw]{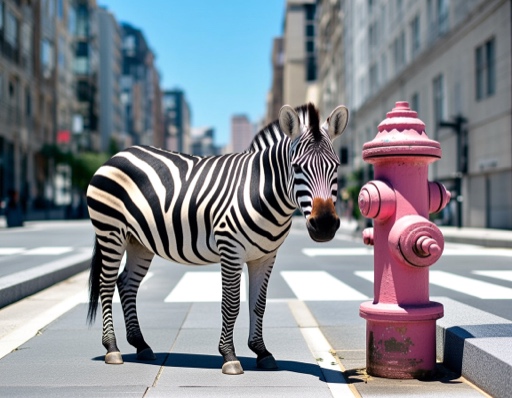} &
  \includegraphics[width=\imgw]{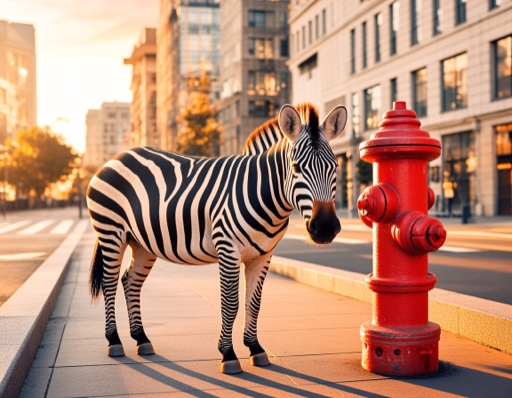}
  \\[-2pt]

  \parbox{\imgw}{\scriptsize\centering \textbf{}} &
  \parbox{\imgw}{\scriptsize\centering \textbf{Fire hydrant to {\tiny(70.8, 87.5, 25.2, 95.2)}}} &
  \parbox{\imgw}{\scriptsize\centering\textbf{
Fire hydrant to}
\fcolorbox{black}{hydrantpink}{}
} & 
  \parbox{\imgw}{\scriptsize\centering \textbf{Warmer color palette}}
  \\[+4pt]

  \includegraphics[width=\imgw]{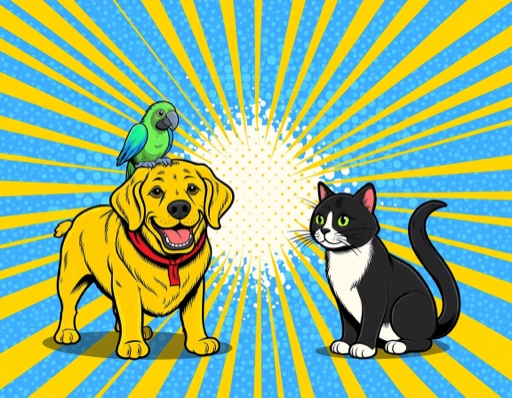} &
  \includegraphics[width=\imgw]{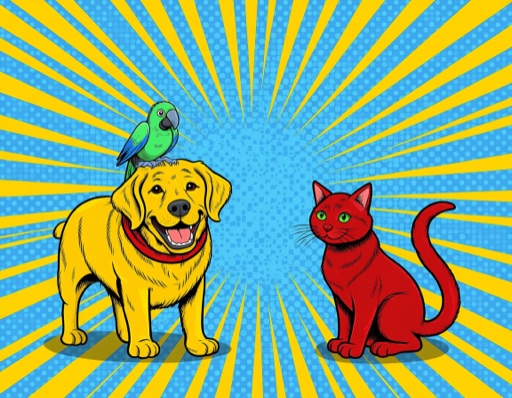} &
  \includegraphics[width=\imgw]{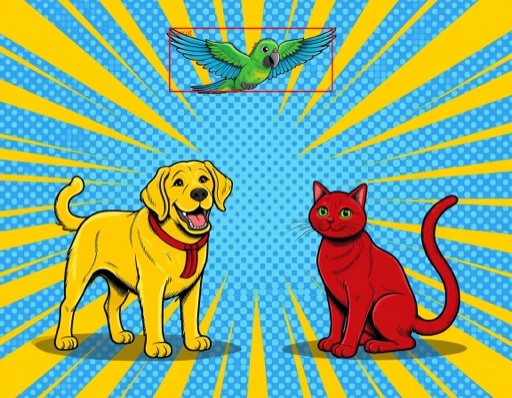} &
  \includegraphics[width=\imgw]{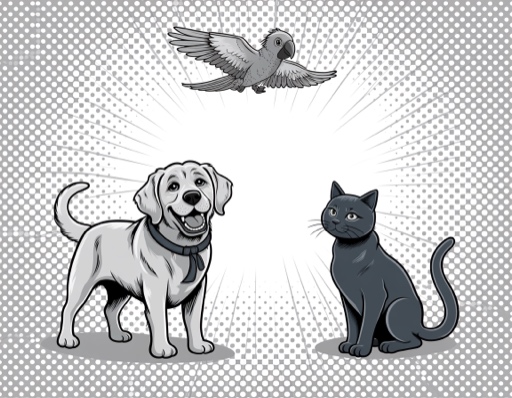}
  \\[-2pt]

  \parbox{\imgw}{\scriptsize\centering \textbf{}} &
  \parbox{\imgw}{\scriptsize\centering\textbf{
Cat to\\}
\fcolorbox{black}{catred}{}
} &
  \parbox{\imgw}{\scriptsize\centering \textbf{Parrot flying at {\tiny(32.3, 66.3, 6.6, 23.5)}}} & 
  \parbox{\imgw}{\scriptsize\centering \textbf{Grayscale}}
  \\[+4pt]



  \includegraphics[width=\imgw]{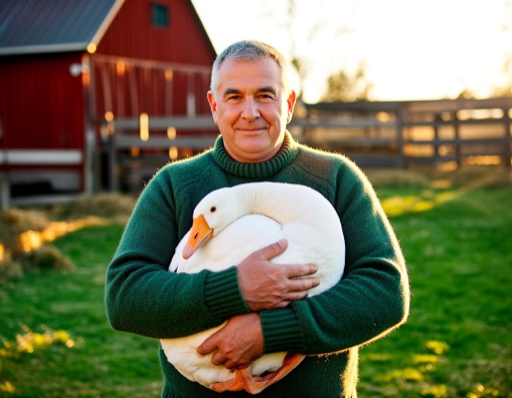} &
  \includegraphics[width=\imgw]{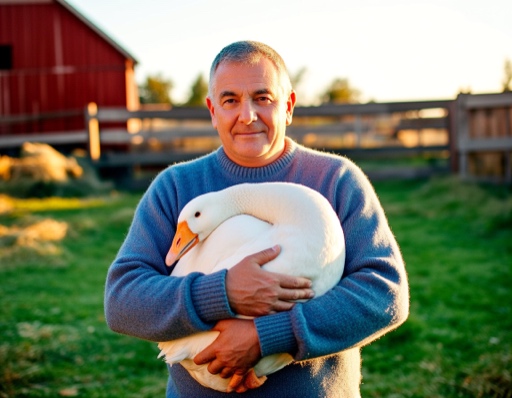} &
  \includegraphics[width=\imgw]{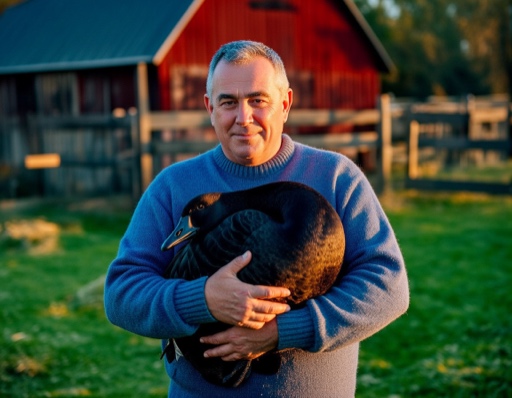} &
  \includegraphics[width=\imgw]{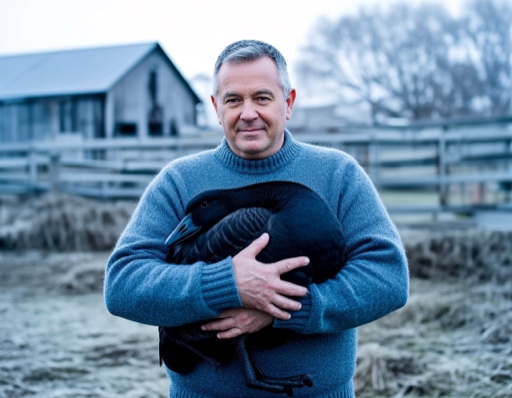}
  \\[-2pt]

  \parbox{\imgw}{\scriptsize\centering \textbf{}} &
  \parbox{\imgw}{\scriptsize\centering\textbf{
Shirt to\\}
\fcolorbox{black}{shirtblue}{}
} &
  \parbox{\imgw}{\scriptsize\centering \textbf{Goose to black}} & 
\parbox{\imgw}{\scriptsize\centering
\textbf{Colder color palette}}
  \\[+2pt]
  
\end{tabular}

\vspace{-3pt}
\caption{\textbf{Disentangled parametric refinement via structured re-generation.}
Each example starts from an image generated from a structured JSON prompt. We then edit only the relevant JSON fields and re-generate using the same random seed. Although the model does not observe the original image, it produces localized changes that follow the modified parameters while preserving the rest of the scene, demonstrating strong parametric disentanglement. Ground-truth bounding boxes are overlaid for visualization.}
\label{fig:disentanglement}
\end{figure}

\setlength{\tabcolsep}{0.5pt}
\renewcommand{\arraystretch}{1.0}

\begin{figure}[t]
\centering
\newcommand{\imgw}{0.19\linewidth} 

\begin{tabular}{@{}ccccc@{}}

  \includegraphics[width=\imgw]{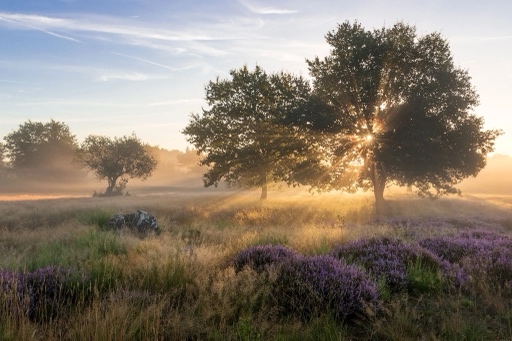} &
  \includegraphics[width=\imgw]{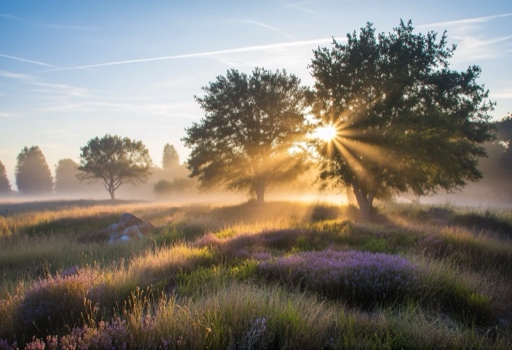} &
  \includegraphics[width=\imgw]{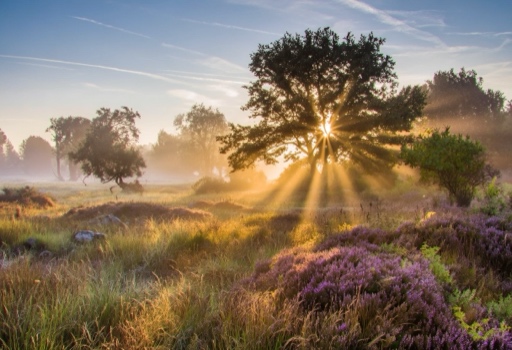} &
  \includegraphics[width=\imgw]{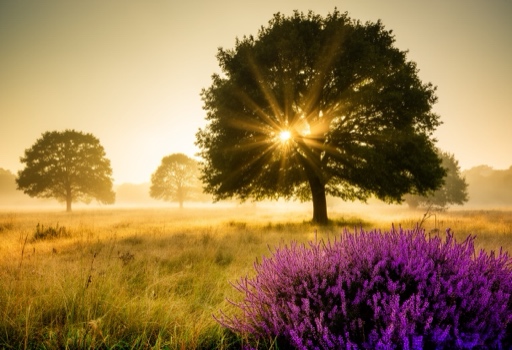} &
  \includegraphics[width=\imgw]{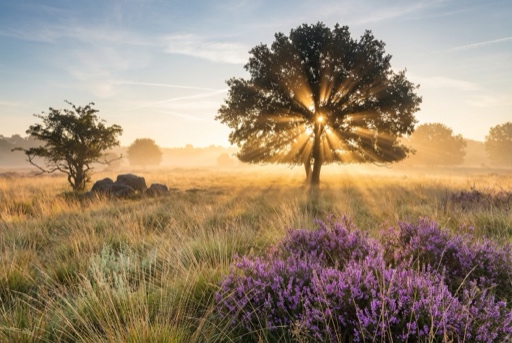}
  \\[-2pt]

  \includegraphics[width=\imgw]{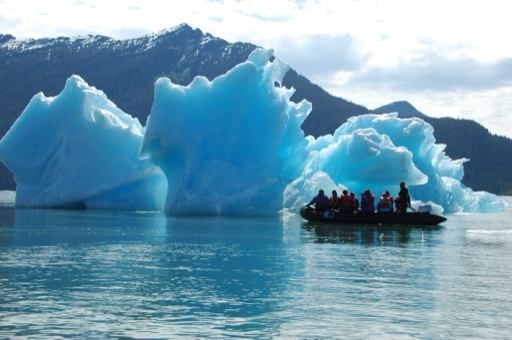} &
  \includegraphics[width=\imgw]{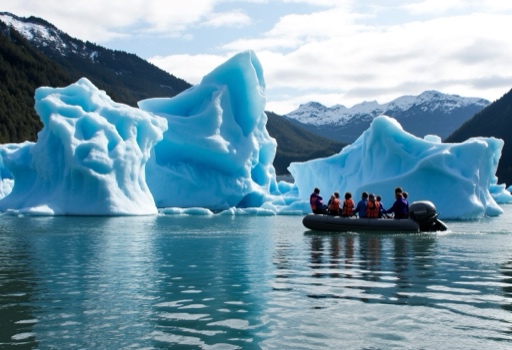} &
  \includegraphics[width=\imgw]{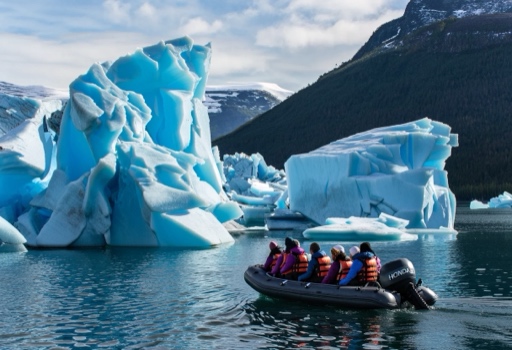} &
  \includegraphics[width=\imgw]{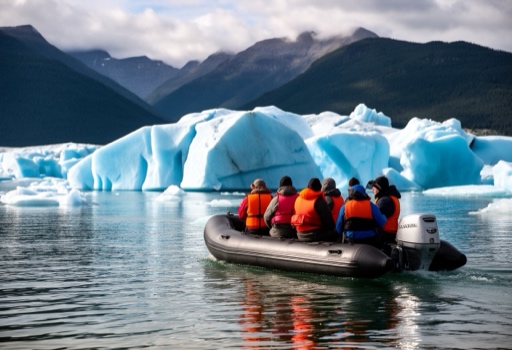} &
  \includegraphics[width=\imgw]{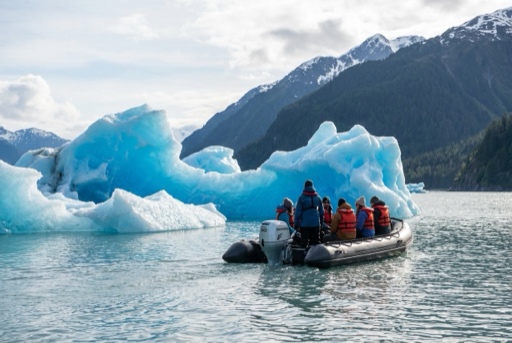}
  \\[-2pt]

  \includegraphics[width=\imgw]{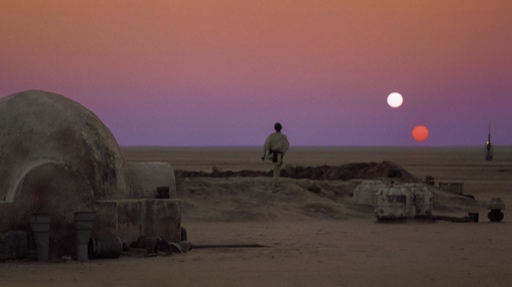} &
  \includegraphics[width=\imgw]{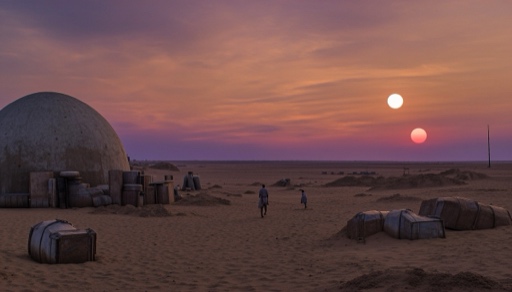} &
  \includegraphics[width=\imgw]{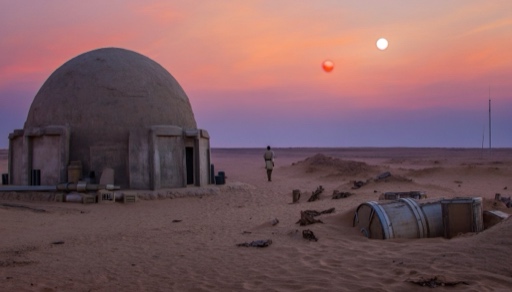} &
  \includegraphics[width=\imgw]{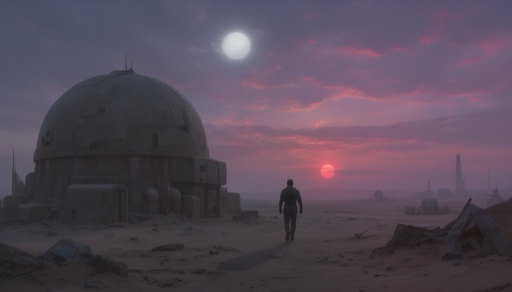} &
  \includegraphics[width=\imgw]{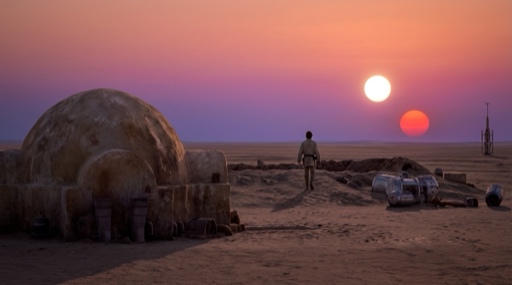}
  \\[-2pt]

  \includegraphics[width=\imgw]{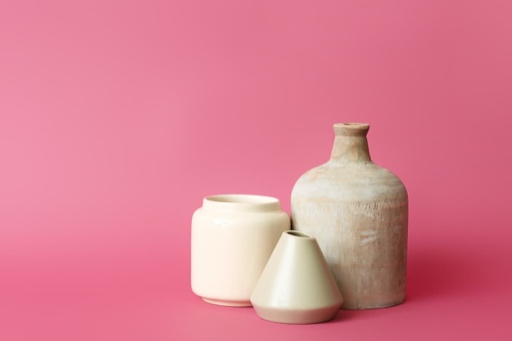} &
  \includegraphics[width=\imgw]{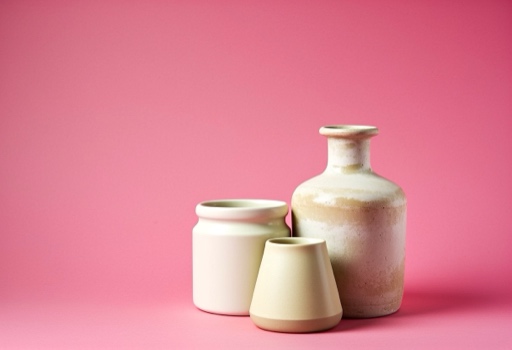} &
  \includegraphics[width=\imgw]{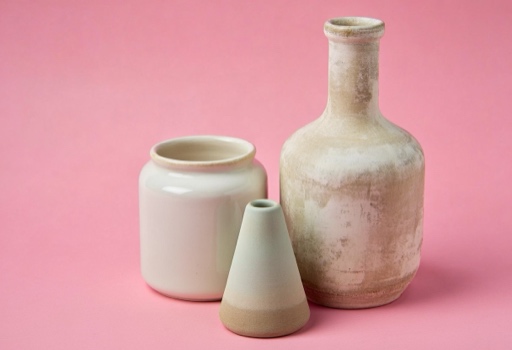} &
  \includegraphics[width=\imgw]{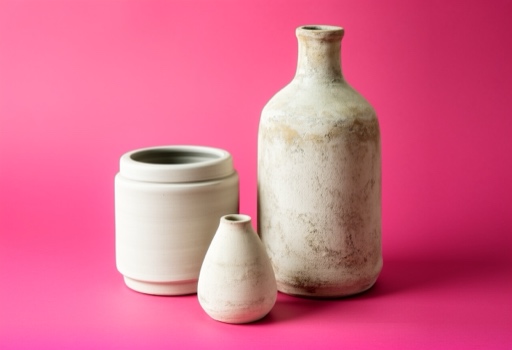} &
  \includegraphics[width=\imgw]{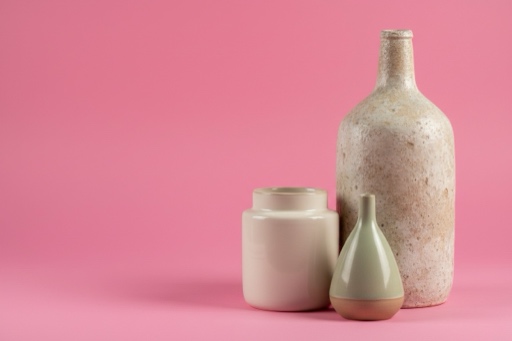}
  \\[-2pt]


  \includegraphics[width=\imgw]{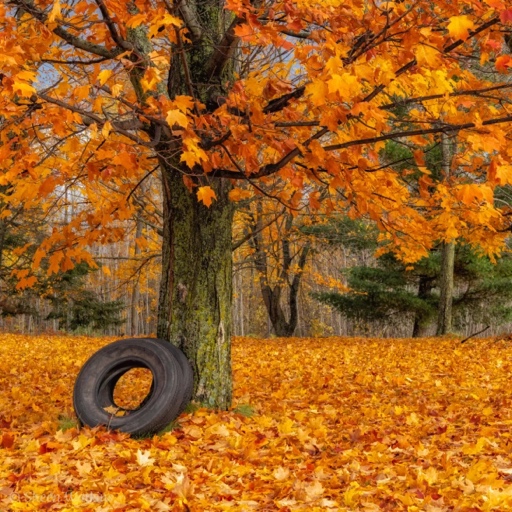} &
  \includegraphics[width=\imgw]{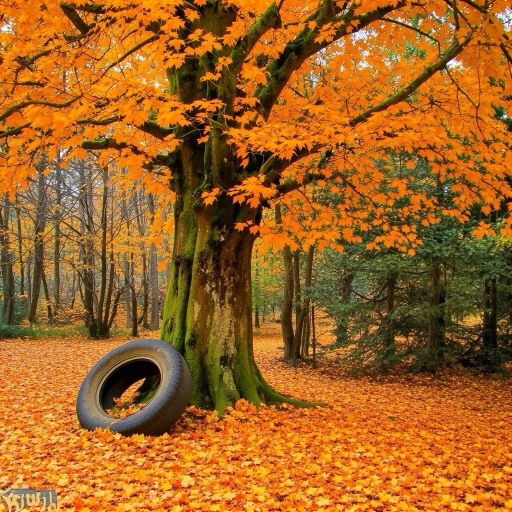} &
  \includegraphics[width=\imgw]{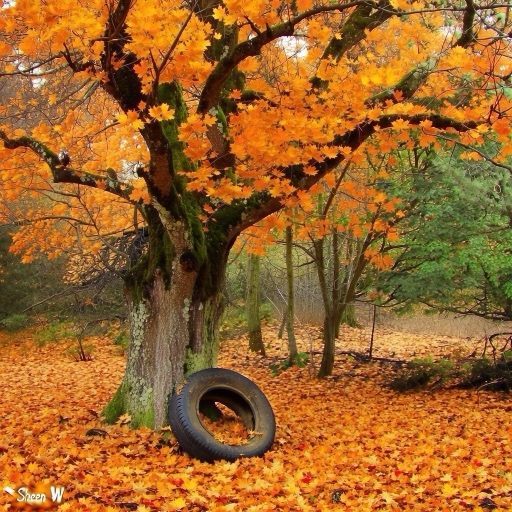} &
  \includegraphics[width=\imgw]{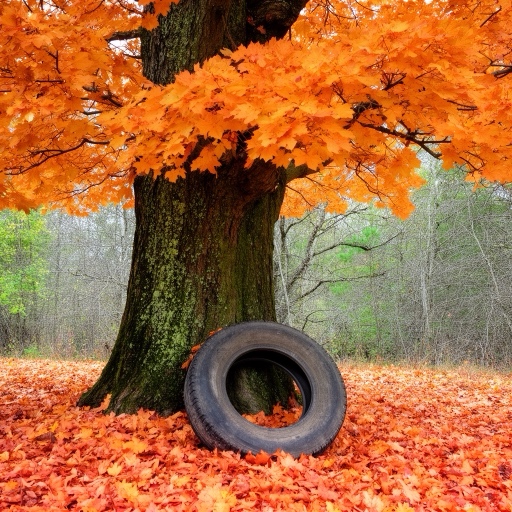} &
  \includegraphics[width=\imgw]{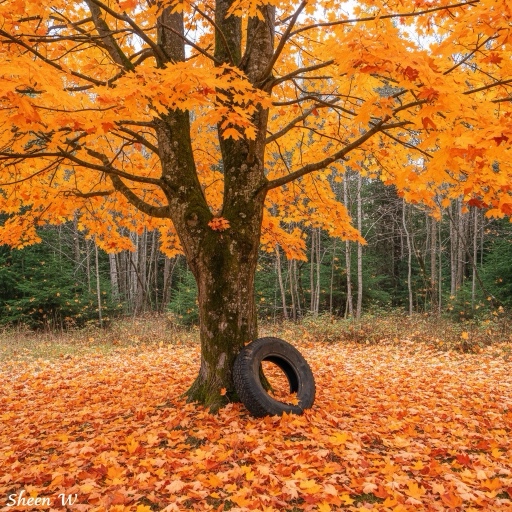}
  \\[-2pt]
  
  \\[-12pt]

  \parbox{\imgw}{\scriptsize\centering \textbf{Original}} &
  \parbox{\imgw}{\scriptsize\centering \textbf{\modelname{} (Ours)}} &
  \parbox{\imgw}{\scriptsize\centering \textbf{FIBO}} &
  \parbox{\imgw}{\scriptsize\centering \textbf{Flux.2}} &
  \parbox{\imgw}{\scriptsize\centering \textbf{NB}}
  \\
\end{tabular}
\vspace{-3pt}
\caption{\textbf{Text-as-a-Bottleneck Reconstruction (TaBR).} 
Starting from the original image (left), a detailed caption is generated and used as input to each model. The resulting reconstructions are compared against the original. BBQ more faithfully preserves scene layout, object relations, and fine-grained attributes than competing state-of-the-art models, demonstrating improved expressiveness.}
\label{fig:tabr}
\end{figure}

\section{Method}\label{sec:method}

We now describe our framework, \modelname{}. Our objective is to adapt a large-scale text-to-image model to accept \emph{numeric} bounding boxes and colors as conditioning inputs, such that the generated image is faithfully aligned with these parametric specifications.

Formally, let $\mathcal{M}$ denote a text-to-image model trained to generate images conditioned on a long structured caption $\mathcal{P}$~\cite{gutflaish2025generating}. 
We extend this model to additionally condition on numeric bounding boxes $\{b_i\}_{i=1}^N$ and colors $\{c_i\}_{i=1}^N$ for each of the $N$ objects in $\mathcal{P}$, producing an image
\[
\mathcal{M}(\mathcal{P}, \{b_i\}_{i=1}^N, \{c_i\}_{i=1}^N)
\]
that is accurately aligned with the specified parameters. Unlike standard text-to-image generation, where spatial and chromatic attributes are described linguistically, bounding boxes and colors in our framework are represented numerically: 
(1)~each bounding box is defined as
$b = (x_0, y_0, x_1, y_1) \in (0,1)^4$, where $(x_0,y_0)$ and $(x_1,y_1)$ are the relative coordinates corresponding to the top-left and bottom-right of the bounding box, and
(2)~each color is defined as an RGB triplet $c \in [0,255]^3$.

In this section, we show that such adaptation is feasible at large scale \emph{without} architectural changes or additional
loss functions, relying solely on dataset augmentation.
In Section~\ref{sec:method:image-to-json}, we describe how we augment structured training captions with numeric bounding boxes and colors $(\{b_i\}_{i=1}^N, \{c_i\}_{i=1}^N)$. Section~\ref{sec:method:bbq} details the training procedure of \modelname{}, including the incorporation of parametric supervision without architectural modifications. Finally, in Section~\ref{sec:method:prompt-to-JSON}, we describe how we bridge the gap between user intent and a valid structured prompt. Specifically, we first present the translation of a short natural-language caption into a full long structured parametric prompt $(\mathcal{P}, \{b_i\}_{i=1}^N, \{c_i\}_{i=1}^N)$, and then describe how users can interactively modify bounding boxes or colors, e.g., by dragging objects or adjusting color values, while maintaining global consistency within the structured representation.

\subsection{Enriching the Training Data with Bounding Boxes and Colors}\label{sec:method:image-to-json}
In \modelname{}, we extend the common practice of synthetic captioning for text-to-image training. Starting from long structured captions~\cite{gutflaish2025generating}, we augment each caption with \emph{numeric} bounding boxes and RGB colors. Although extracting such parameters is well studied in vision and graphics, we find that general-purpose LLM/VLM systems (e.g., Gemini~2.5~\cite{comanici2025gemini}) are not sufficiently reliable for high-precision outputs.
Therefore, for each image we first generate a FIBO-style structured caption, following~\cite{gutflaish2025generating}. For every object mentioned in the caption, we extract its bounding box from grounded SAM2~\cite{ravi2024sam2}, estimate relative depth using Depth Anything V2~\cite{depth_anything_v2}, and obtain dominant object colors using Pylette~\cite{pylette2025}.
We replace semantic location and qualitative color terms with explicit bounding box coordinates and RGB triplets. Finally, a global RGB palette from Pylette is added to capture the overall color scheme. This automated extraction provides the precise parametric grounding required to align numeric tokens with visual synthesis.

\setlength{\tabcolsep}{0.5pt}
\renewcommand{\arraystretch}{1.0}

\begin{figure}[t]
\centering
\newcommand{\imgw}{0.19\linewidth} 

\begin{tabular}{@{}ccccc@{}}
  \includegraphics[width=\imgw]{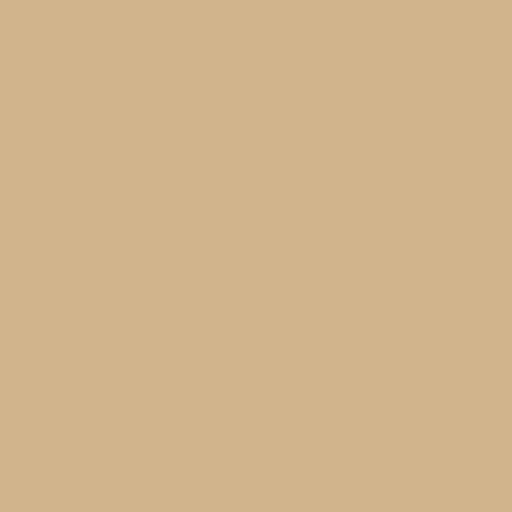} &
  \includegraphics[width=\imgw]{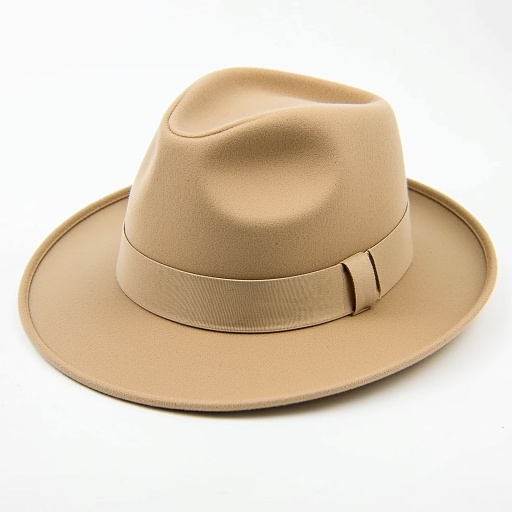}&
  \includegraphics[width=\imgw]{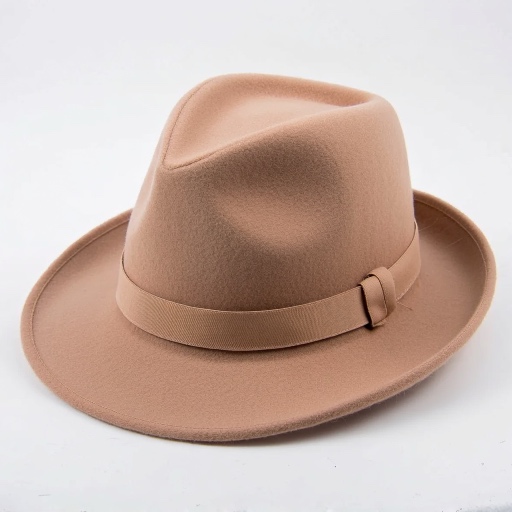} &
  \includegraphics[width=\imgw]{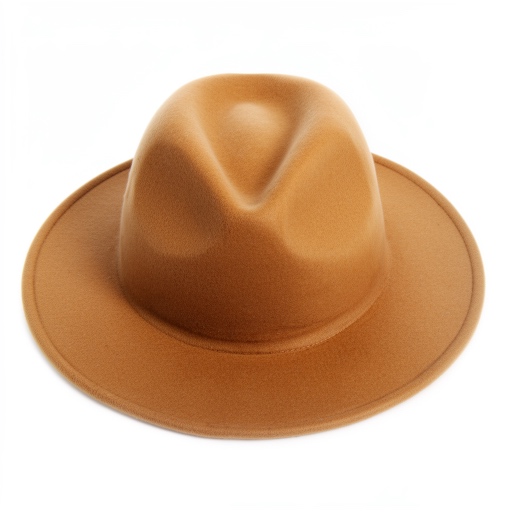} &
  \includegraphics[width=\imgw]{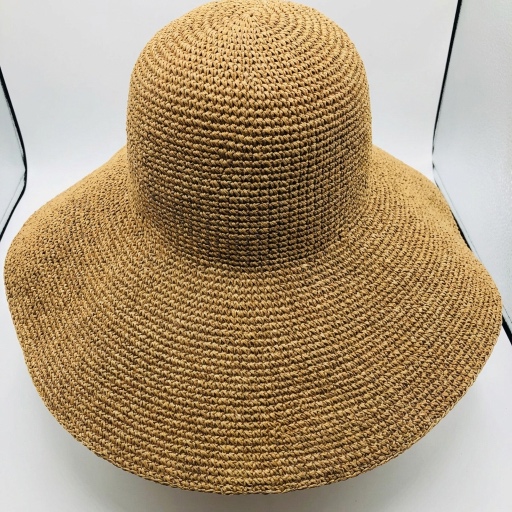}
  \\[-2pt]

  \includegraphics[width=\imgw]{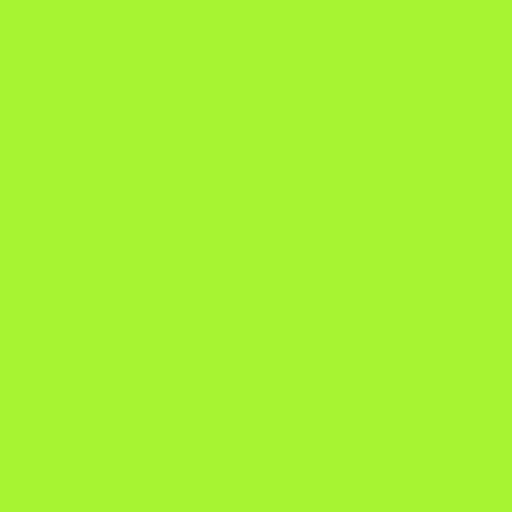} &
  \includegraphics[width=\imgw]{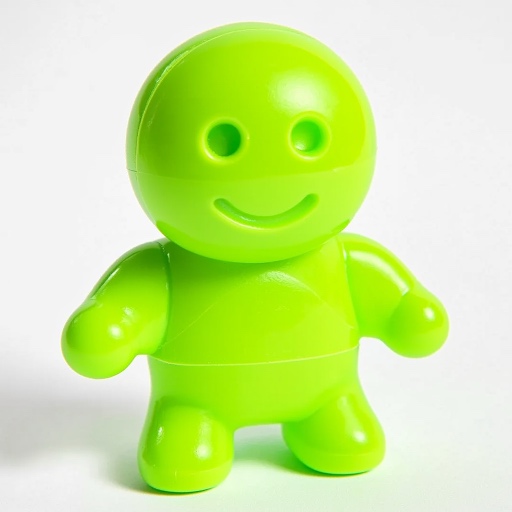}&
  \includegraphics[width=\imgw]{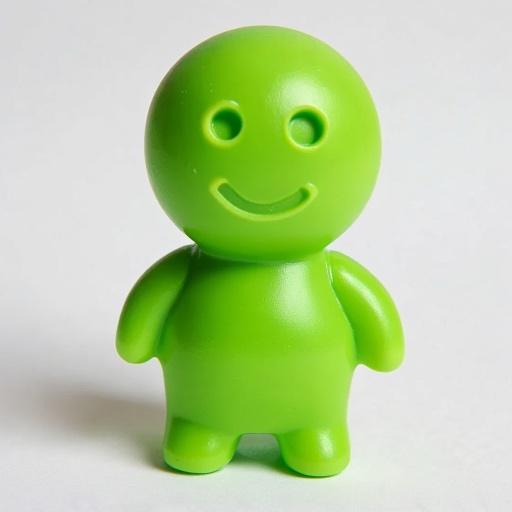} &
  \includegraphics[width=\imgw]{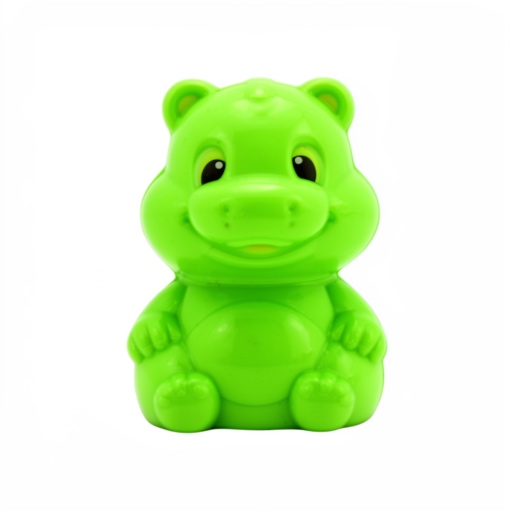} &
  \includegraphics[width=\imgw]{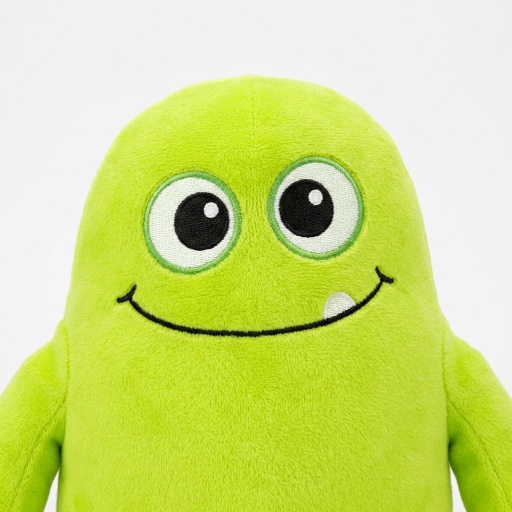}
  \\[-2pt]

  \includegraphics[width=\imgw]{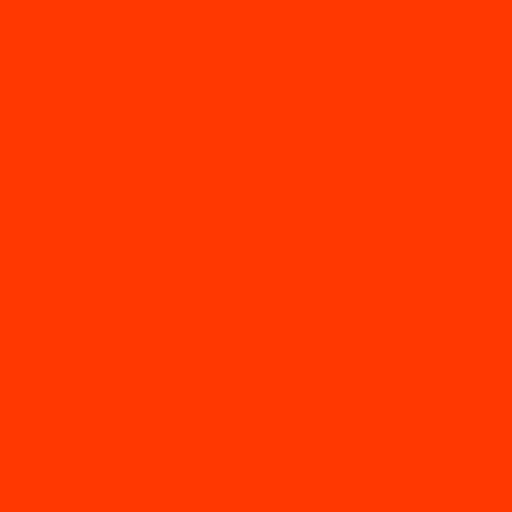} &
  \includegraphics[width=\imgw]{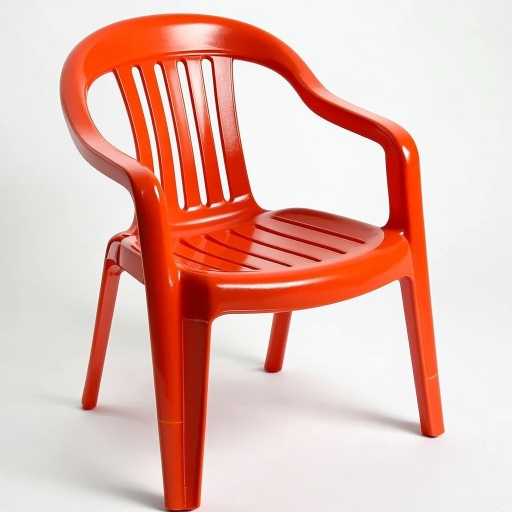}&
  \includegraphics[width=\imgw]{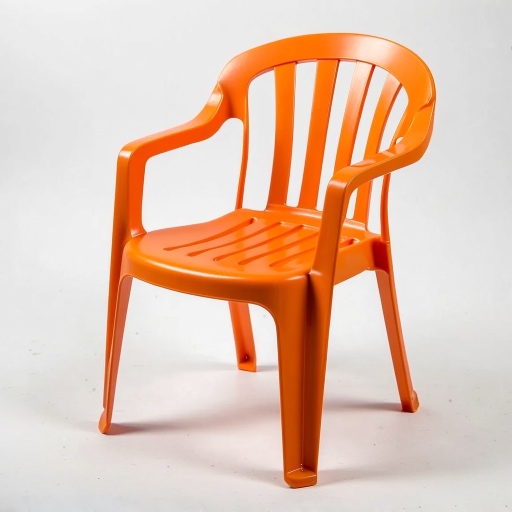} &
  \includegraphics[width=\imgw]{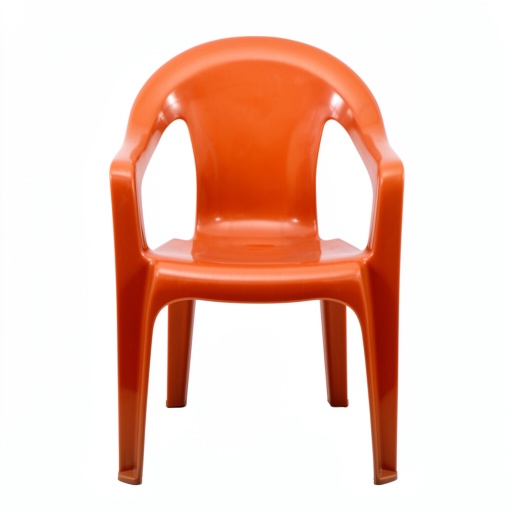} &
  \includegraphics[width=\imgw]{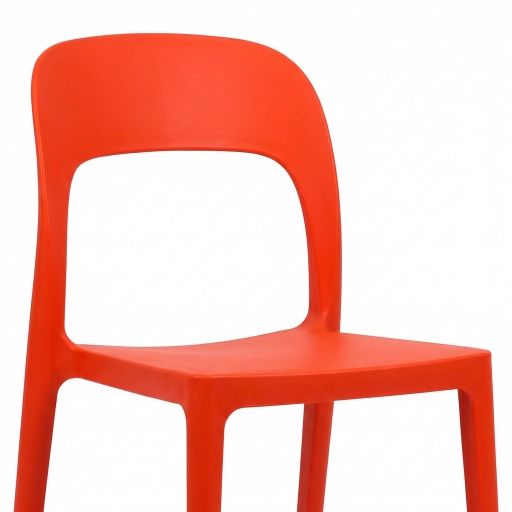}
  \\[-2pt]


  \includegraphics[width=\imgw]{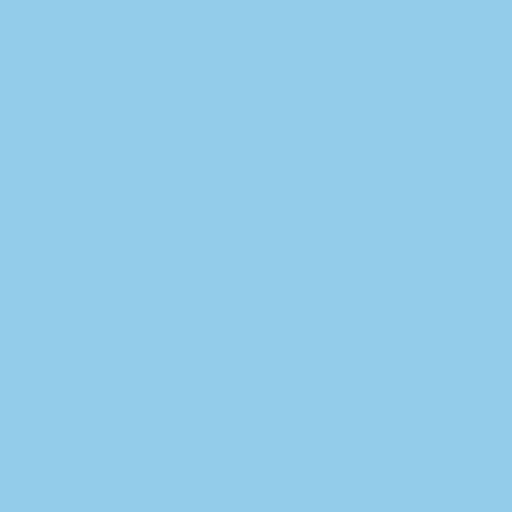} &
  \includegraphics[width=\imgw]{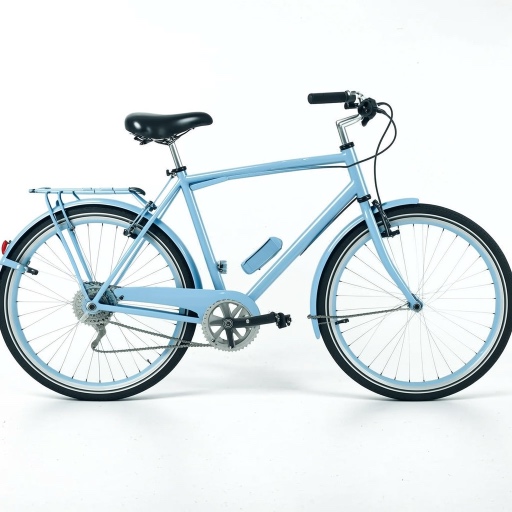}&
  \includegraphics[width=\imgw]{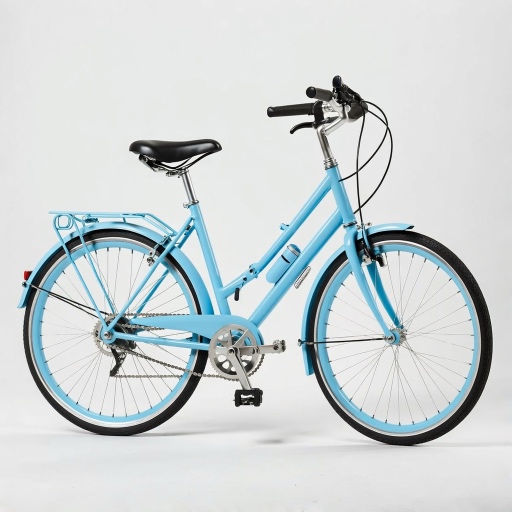} &
  \includegraphics[width=\imgw]{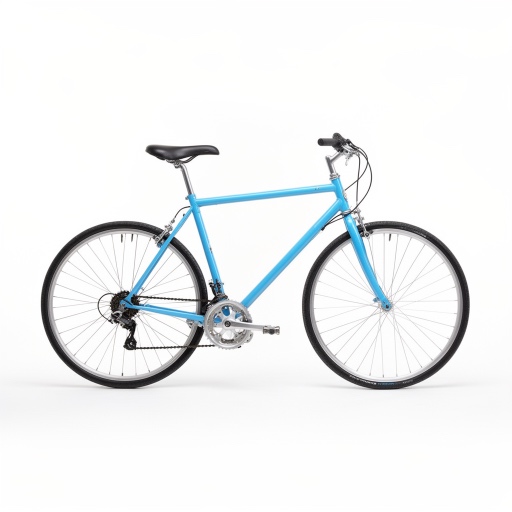} &
  \includegraphics[width=\imgw]{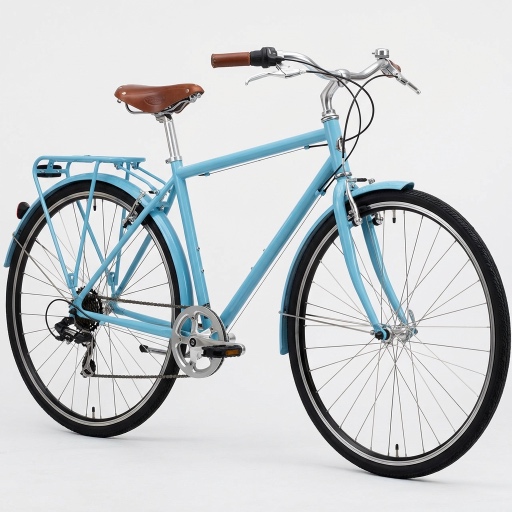}
  \\[-2pt]

  \parbox{\imgw}{\scriptsize\centering \textbf{Target Color}} &
  \parbox{\imgw}{\scriptsize\centering \textbf{\modelname{} (Ours)}} &
  \parbox{\imgw}{\scriptsize\centering \textbf{FIBO}} &
  \parbox{\imgw}{\scriptsize\centering \textbf{Flux.2}} &
  \parbox{\imgw}{\scriptsize\centering \textbf{NB}}
  \\
\end{tabular}
\vspace{-3pt}
\caption{\textbf{Color-conditioning accuracy.} 
Each example shows the target color (left) and images generated by different models when conditioned on the same object and exact RGB value. \modelname achieves high chromatic fidelity to the target color and produces competitive results compared to state-of-the-art text-to-image models under identical color-conditioning prompts.}
\label{fig:colors}
\end{figure}

\subsection{\modelname: Large-Scale Training to Control Bounding Boxes and Qolors}\label{sec:method:bbq}

Unlike prior approaches that introduce new architectures, loss functions, or extended inference procedures to enable parametric control, we show that strong bounding-box grounding can be achieved by large-scale training on enriched captions alone. We initialize from the $8$B-parameter FIBO backbone~\cite{gutflaish2025generating}, which is designed to process long structured captions, and continue training on 25M images paired with our parametric captions. 

We train the model following FIBO's hyperparameters~\cite{gutflaish2025generating}, using the AdamW optimizer~\cite{loshchilov2017decoupled} with weight decay of $1\times10^{-4}$, $\beta_1=0.9$, $\beta_2=0.999$, and $\epsilon=1\times10^{-15}$. The learning rate is set to $1\times10^{-4}$ with a constant schedule and a warmup of $10$K steps. Training follows the flow-matching formulation~\cite{lipman2023flowmatchinggenerativemodeling}, with a logit-normal noise schedule combined with resolution-dependent timestep shifting~\cite{esser2024scaling}.
The model was trained for 80,000 steps with an effective batch size of 512 in resolution $1024^2$.
Post-training, we perform aesthetic finetuning with 3,000 hand-picked images, followed by DPO training~\cite{wallace2024diffusion} with dynamic beta~\cite{liu2025improving} to improve text rendering.

As shown in Figure~\ref{fig:teaser}, \modelname{} adapts effectively to the new conditioning format and follows numeric inputs with high fidelity. Furthermore, Figure~\ref{fig:disentanglement} demonstrates that \modelname{} preserves FIBO’s native disentanglement: using the same random seed, we modify only the relevant fields in the structured JSON and re-generate the image, resulting in targeted changes to the specified attribute while the rest of the scene remains largely unchanged.

\begin{figure}[t]
\centering
\setlength{\tabcolsep}{4pt}
\renewcommand{\arraystretch}{1.0}
\newcommand{\imgw}{0.235\linewidth} 

\parbox{0.98\linewidth}{\scriptsize\itshape
``A knight located at (top left: (27.2, 36.3), bottom right: (54.8, 98))
is going towards a dragon at (top left: (63.1, 14.7), bottom right: (77.1, 29.1)).''
}\vspace{2pt}

\begin{tabular}{@{}c c c@{}}
\includegraphics[width=\imgw]{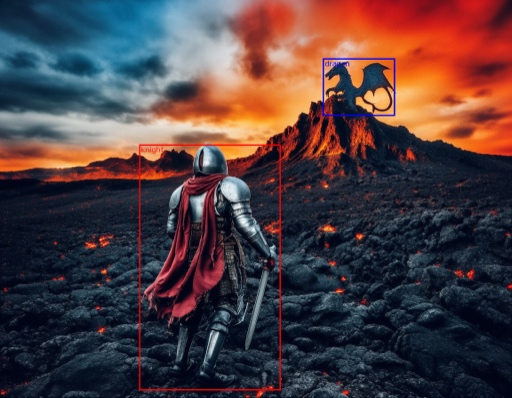} &
\includegraphics[width=\imgw]{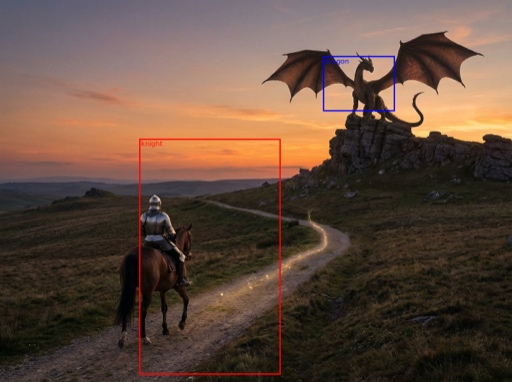} &
\includegraphics[width=\imgw]{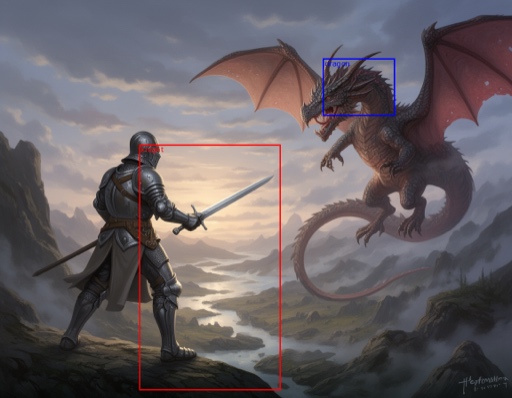} \\
\scriptsize\bfseries BBQ (ours) & \scriptsize\bfseries NB & \scriptsize\bfseries Flux.2 \\
\end{tabular}

\vspace{6pt}

\parbox{0.98\linewidth}{\scriptsize\itshape
``Three glass bottles standing in a row: a red bottle at (top left: (12.5, 30), bottom right: (27.5, 80)),
a green bottle at (top left: (42.5, 30.0), bottom right: (57.5, 80.0)),
and a blue bottle at (top left: (72.6, 30.0), bottom right: (87.6, 80.0)).''
}\vspace{2pt}

\begin{tabular}{@{}c c c@{}}
\includegraphics[width=\imgw]{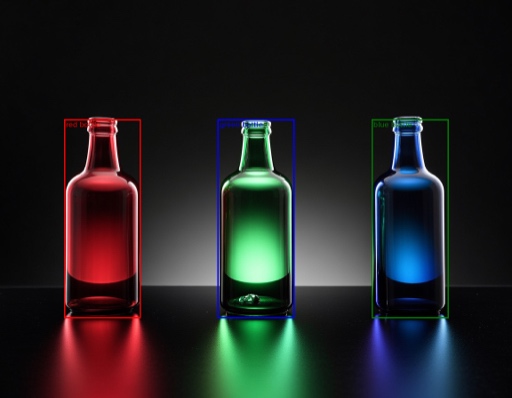} &
\includegraphics[width=\imgw]{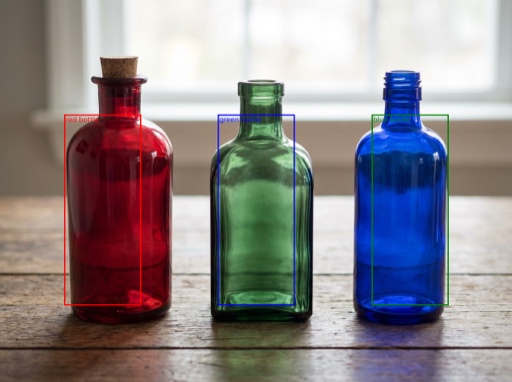} &
\includegraphics[width=\imgw]{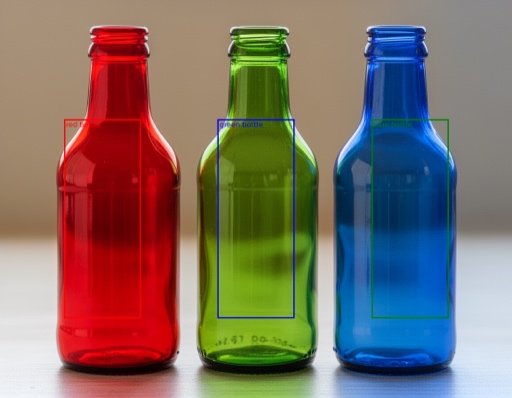} \\
\scriptsize\bfseries BBQ (ours) & \scriptsize\bfseries NB & \scriptsize\bfseries Flux.2 \\
\end{tabular}

\parbox{0.98\linewidth}{\scriptsize\itshape
``A monkey at ("top left": (16.5, 59.8), "bottom right": (35, 85)) is going towards a zebra at ("top left": (54.3, 17.1), "bottom right": (92.3, 89.7))''
}\vspace{2pt}

\begin{tabular}{@{}c c c@{}}
\includegraphics[width=\imgw]{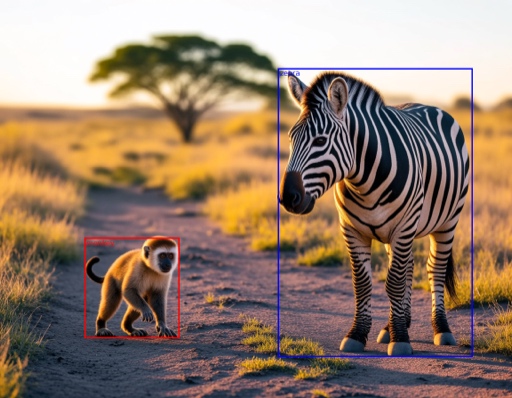} &
\includegraphics[width=\imgw]{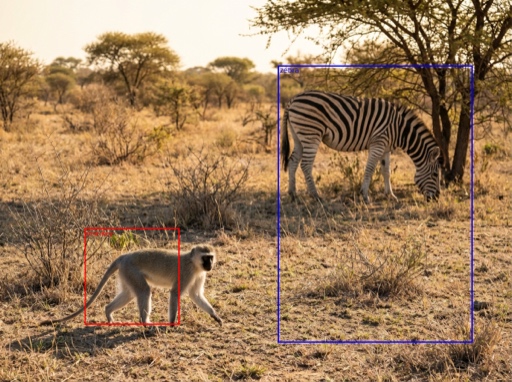} &
\includegraphics[width=\imgw]{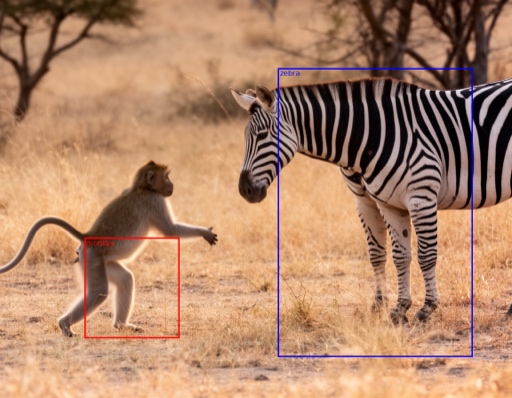} \\
\scriptsize\bfseries BBQ (ours) & \scriptsize\bfseries NB & \scriptsize\bfseries Flux.2 \\
\end{tabular}

\caption{\textbf{Bounding-box accuracy.} 
We compare \modelname with Nano Banana Pro and Flux.2 Pro on prompts that include explicit numeric bounding-box specifications (overlaid on the images). While the baseline models often struggle to consistently follow these spatial constraints, \modelname  reliably places objects within the specified boxes.
}
\label{fig:boxes}
\end{figure}

\subsection{The Parametric Bridge: From Short Captions to Long, Structured, Parametric Prompts}\label{sec:method:prompt-to-JSON}

The trained model enables new forms of user interaction, including object dragging, resizing, and recoloring. However, building a complete end-to-end system introduces two key challenges. First, when a user edits a bounding box, the system must preserve global coherence and avoid breaking the composition. For example, if two people are hugging and the user separates their boxes, the underlying action must necessarily change. Second, for generation from scratch, a short natural-language prompt must be expanded into a full structured caption with a plausible composition, now including explicit bounding boxes and colors. While \modelname{} provides unprecedented precision through its parametric schema, manually authoring JSON prompts with exact RGB triplets and normalized bounding box coordinates is impractical for human users.

To address these inference-time gaps, we fine-tune \mbox{Qwen-3~VL} 4B~\cite{Qwen3-VL} to serve as an inference-time \emph{bridge} that translates natural-language intent into the parametric language consumed by the generator. We train on synthetically generated short prompts and editing instructions, using the same structured schema employed for FIBO BBQ. Training is performed on $8\times$H100 with a total of $3$B tokens. To improve robustness, we decouple image-conditioned and text-only tasks during training and repeat each with different seeds, then final weights are produced via model merging~\cite{yang2024model}.

The VLM operates in three modes: (1) \emph{Generate}, which expands a brief prompt into a complete parametric JSON; (2) \emph{Refine}, which edits an existing JSON in response to textual instructions (e.g., shifting bounding boxes or adjusting colors) while maintaining internal consistency; and (3) \emph{Inspire}, which extracts a parametric description from a reference image to serve as a template for generation and editing. In practice, we find that state-of-the-art VLMs such as Gemini~2.5~\cite{comanici2025gemini} can also serve as an effective inference-time bridge. 
The workflow of \modelname is described in Figure~\ref{fig:workflow}.

\begin{table}[t]
\centering
\caption{\textbf{Text-as-a-Bottleneck Reconstruction (TaBR).} Win rate is computed as the fraction of images where BBQ is preferred over the competing model among decisive comparisons (ties ignored). Confidence intervals correspond to 95\% Wilson score intervals. BBQ outperforms all evaluated baselines across all comparisons.}
\label{tab:tabr}
\begin{tabular}{lcc}
\toprule
Model (vs.\ BBQ) \ \ \ & BBQ win rate$\uparrow$ & 95\% CI \\
\midrule
Nano Banana Pro & 65.2\% & [50.8, 77.3] \\
FIBO & 76.1\% & [62.1, 86.1] \\
FLUX.2 Pro & 93.3\% & [82.1, 97.7] \\
\bottomrule
\end{tabular}
\end{table}

\section{Experiments}

In this section, we present a comprehensive evaluation of \modelname, comparing it to existing state-of-the-art models.
Our experiments are designed to isolate three complementary properties: (1)~expressiveness, (2)~spatial accuracy under numeric box constraints, and (3)~color fidelity under explicit RGB specification.
Evaluation methods are described in Section~\ref{sec:experiments:metrics}, qualitative results are provided in Section~\ref{sec:experiments:qualitative}, and quantitative results are discussed in Section~\ref{sec:experiments:quantitative}.

\subsection{Evaluation Metrics}\label{sec:experiments:metrics}

We evaluate \modelname{} using three complementary metrics that capture different aspects of controlled image synthesis: 
(1)~\emph{Text-as-a-Bottleneck Reconstruction (TaBR)}~\cite{gutflaish2025generating} measures overall expressiveness via caption$\rightarrow$generation$\rightarrow$reconstruction, 
(2)~\emph{Bounding-box accuracy} measures spatial grounding under box-conditioned prompts, using COCO with YOLO-based detection~\cite{lin2014microsoft,Jocher_Ultralytics_YOLO_2023} and LVIS with box-conditioned zero-shot grounding~\cite{gupta2019lvis}, and  
(3)~\emph{Color accuracy} measures parametric color fidelity by clustering generated pixels in CIELab space using K-means and reporting perceptual color differences via CIEDE2000 ($\Delta E_{00}$) and the $a$--$b$ chroma distance.
For TaBR and color accuracy, we compare \modelname{} against state-of-the-art text-to-image baselines (FIBO, Nano Banana Pro, and Flux.2 Pro). 
For bounding-box accuracy, we additionally compare against InstanceDiffusion~\cite{wang2024instancediffusion} and GLIGEN~\cite{li2023gligen}, widely used  box-grounded generation methods.

\begin{table*}[t]
\caption{\textbf{Bounding-box alignment under box-conditioned generation on COCO and LVIS.}
We follow the InstanceDiffusion evaluation protocol using YOLO-based detection; the upper bound corresponds to detector performance on real images.
Across both datasets, \modelname consistently outperforms strong text-to-image baselines (Nano Banana Pro and Flux.2 Pro) and GLIGEN, while trailing the specialized InstanceDiffusion approach.
Importantly, \modelname achieves this without architectural modifications or grounding-specific components and is trained for high-fidelity image synthesis, providing strong spatial control within a general large-scale  and disentangle model, also allowing intuitive refinement.
Best results are in \textbf{bold}; second best are \underline{underlined}.}
\label{tab:box_alignment_coco_lvis}

\centering
\small
\setlength{\tabcolsep}{4pt}
\begin{tabular}{l|ccc|cccccccc}
\hline
& \multicolumn{3}{c|}{\textbf{COCO}} & \multicolumn{8}{c}{\textbf{LVIS}} \\
\textbf{Method}
& $\mathbf{AP}$ & $\mathbf{AP_{50}}$ & $\mathbf{AR}$
& \textbf{AP} & $\mathbf{AP_{50}}$ & $\mathbf{AP_{s}}$ & $\mathbf{AP_{m}}$ & $\mathbf{AP_{l}}$
& $\mathbf{AP_{r}}$ & $\mathbf{AP_{c}}$ & $\mathbf{AP_{f}}$ \\
\hline
\rowcolor{gray!15}
Upper bound (real images)
& 50.2 & 66.7 & 61.0
& 44.6 & 57.7 & 33.2 & 55.0 & 66.1 & 31.4 & 44.5 & 50.5 \\
\hline
Flux.2 Pro
& 3.5 & 8.7 & 4.6
& 2.1 & 4.8 & 0.1 & 0.8 & 7.3 & 0.1 & 0.2 & 2.4 \\
Nano Banana Pro
& 5 & 11.3 & 5.5
& 4.1 & 10.6 & 0.1 & 1 & 15.9 & 3.3 & 4 & 3.9 \\
GLIGEN \cite{li2023gligen}
& 19.6 & 35.0 & 30.7
& 9.9 & 9.5 & \underline{1.6} & 10.5 & 31.1 & 7.4 & 10.0 & 10.9 \\
BBQ (ours)
& \underline{28.6} & \underline{40.9} & \underline{38.2}
& \underline{13.1} & \underline{20.1} & \underline{1.6} & \underline{13.9} & \underline{37.4} & \textbf{13.1} & \underline{14.0} & \underline{12.7} \\
InstanceDiffusion \cite{wang2024instancediffusion}
& \textbf{38.8} & \textbf{55.4} & \textbf{52.9}
& \textbf{17.9} & \textbf{25.5} & \textbf{5.5} & \textbf{24.2} & \textbf{45.0} & \underline{12.7} & \textbf{18.7} & \textbf{19.3} \\
\hline
\end{tabular}
\end{table*}
\begin{table*}[t]
\centering
\caption{\textbf{Color fidelity comparison.} 
$\Delta E_{00}$ (CIEDE2000) measures perceptual color difference, while $a$--$b$ distance
captures chromaticity (hue and saturation) independently of lightness.
We report mean, median, and 90th percentile (p90), where lower values indicate better color accuracy.. 
Across both $K=5$ and $K=8$, \modelname{} achieves the lowest $a$--$b$ errors in all statistics,
indicating the most accurate chromaticity control and fewer severe failures, while remaining
competitive under $\Delta E_{00}$ that penalizes lightness differences.
Best results are in \textbf{bold}
and second-best are \underline{underlined} (computed per $K$).
}
\label{tab:color_fidelity}
\begin{tabular}{c|lcccccc}
\toprule
~~$K$~~ & ~~ Model &
$a$--$b$ Mean$\downarrow$ \ \ &
$a$--$b$ Median$\downarrow$  \ \ &
$a$--$b$ p90$\downarrow$  \ \ &
$\Delta E_{00}$ Mean$\downarrow$  \ \ &
$\Delta E_{00}$ Median$\downarrow$  \ \ &
$\Delta E_{00}$ p90$\downarrow$  \ \ \\
\midrule
\multirow{4}{*}{5} & ~~BBQ (ours)
& \textbf{7.16} & \textbf{6.67} & \textbf{13.30}
& \textbf{5.93} & \underline{5.76} & \textbf{9.62} \\
& ~~Nano Banana Pro
& 10.91 & \underline{7.52} & 20.80
& \underline{6.50} & \textbf{5.72} & 11.00 \\
& ~~FLUX.2 Pro
& \underline{10.07} & 8.16 & \underline{19.30}
& 6.64 & 6.12 & \underline{10.17} \\
& ~~FIBO
& 10.32 & 9.44 & 20.36
& 6.74 & 7.32 & 10.52 \\
\midrule
\multirow{4}{*}{8} & ~~BBQ (ours)
& \textbf{7.48} & \textbf{6.23} & \textbf{14.33}
& \underline{5.74} & \underline{5.07} & \textbf{9.89} \\
& ~~Nano Banana Pro
& 10.64 & \underline{6.87} & 21.05
& 5.91 & \textbf{4.82} & 10.80 \\
& ~~FLUX.2 Pro
& \underline{9.50} & 8.26 & \underline{17.98}
& \textbf{5.67} & 5.42 & \underline{9.99} \\
& ~~FIBO
& 11.07 & 9.62 & 20.52
& 6.99 & 5.90 & 11.03 \\
\bottomrule
\end{tabular}
\end{table*}

\paragraph{Text-as-a-Bottleneck.}
TaBR \cite{gutflaish2025generating} measures the overall expressive power by anchoring the evaluation in images rather than subjective text reasoning. Following FIBO, we begin with a real image, produce a detailed caption using a VLM, and then regenerate the image from this caption alone. Annotators are then presented with the original image alongside two reconstructions from different models and asked: ``Which image is more similar to the original?'' Like in FIBO, we perform this measurement on a test-set of 60 image that are not part of our training data.

For \modelname and FIBO we utilize their native structured schemas for compatibility, while for Nano Banana Pro and Flux.2 Pro, we report the best result among three methods: (a)~\modelname parametric captions, (b)~FIBO long structured captions, and (c)~detailed free-text descriptions including precise Hex codes for object colors. To avoid evaluation bias from the captioning pipeline, we use a \emph{neutral} VLM that is independent of \modelname{} and the data preparation used for FIBO.

\paragraph{YOLO- and LVIS-based scores.}
We follow the evaluation protocol of InstanceDiffusion~\cite{wang2024instancediffusion} to assess spatial alignment between generated images and input bounding boxes.
A pretrained object detector is applied to the generated images, and the predicted boxes are compared against the input box coordinates.
For COCO evaluation, we use YOLOv8 and report $AP$, $AP_{50}$, and $AR$ on COCO2017-val.
For large-vocabulary evaluation, we follow the LVIS protocol using a ViTDet-L detector.

\paragraph{Color-conditioning accuracy.}
To evaluating color fidelity we wish to isolate the specific object and remove noise from other parts of the image. Therefore, we generated 200 images depicting single objects on white background, where each object was assigned a specific target RGB color in the prompt. For evaluation, we extract object pixels by masking out the white background using foreground segmentation, and then apply K-means clustering (with $K = 5$ and $K = 8$) in CIELab color space on the extracted object pixels to identify the dominant color palette. Clusters representing less than 5\% of object pixels are filtered out. Among the remaining clusters, we select the one with the minimum distance to the target color. We report two distance metrics: $\Delta E_{00}$ (CIEDE2000),
which measures perceptual color difference, and Euclidean distance in the a-b
chromaticity plane, which isolates hue and saturation differences independently of light. For both metrics, we report mean, median, and 90th percentile (p90) statistics, where p90 captures tail behavior and robustness to difficult cases that may not be reflected by central tendency alone. Like in TaBR, \modelname and FIBO utilize their native structured schemas for compatibility, where for FIBO we ask the VLM to choose the name of the color that best describes the RGB.  For Flux.2 Pro we follow the prompting guide~\cite{bfl_flux2_prompting_guide} and for Nano Banana Pro we've found that the best results are achieved with the same prompts as Flux.

\subsection{Qualitative Results}\label{sec:experiments:qualitative}

In Figure~\ref{fig:tabr}, we present TaBR reconstructions, where \modelname{} faithfully preserves the original pose, object relationships, and overall scene layout.
Figure~\ref{fig:colors} illustrates \modelname{}’s ability to follow explicit numeric color specifications: when conditioned on exact RGB values, the model produces visually accurate object colors and remains competitive with state-of-the-art baselines.
Figure~\ref{fig:boxes} provides qualitative comparisons for bounding-box grounding against strong general-purpose models, Nano Banana Pro and Flux.2 Pro. While these baselines often struggle to satisfy explicit numeric box constraints, \modelname{} consistently aligns object placement with the specified regions, motivating our subsequent quantitative comparison against dedicated layout-aware approaches such as InstanceDiffusion and GLIGEN.
Additional results are presented in Figure~\ref{fig:refinement_example} demonstrating the effectiveness of our approach.

\subsection{Quantitative Results}\label{sec:experiments:quantitative}

\paragraph{Text-as-a-Bottleneck.}
Table~\ref{tab:tabr} reports image-level pairwise preference results for the TaBR evaluation. We report the win rate of \modelname as the fraction of images where it is preferred over the competing model among decisive outcomes (ties ignored), together with 95\% Wilson score confidence intervals. As shown in the table, \modelname consistently outperforms its predecessor FIBO as well as state-of-the-art general-purpose text-to-image models, including Nano Banana Pro and Flux.2 Pro, demonstrating that incorporating explicit numeric parameters improves reconstruction fidelity without sacrificing global coherence.

\paragraph{Bounding-box accuracy.}
Table~\ref{tab:box_alignment_coco_lvis} evaluates spatial grounding under box-conditioned prompts on COCO and LVIS. 
Across both datasets, \modelname consistently outperforms strong text-to-image baselines such as Nano Banana Pro and Flux.2 Pro, as well as the dedicated grounding model GLIGEN, while trailing the current state-of-the-art InstanceDiffusion. These results position \modelname as a strong non-specialized alternative for box-conditioned generation.
Unlike InstanceDiffusion and GLIGEN, which rely on grounding-specific architectural modifications or inference-time alignment mechanisms,
\modelname is trained at a substantially larger scale for general high-fidelity image synthesis, achieving strong bounding-box alignment without sacrificing expressiveness, inference time or requiring specialized components. Furthermore, unlike InstanceDiffusion, \modelname{} exhibits native disentanglement that enables intuitive parametric refinement, as illustrated in Fig.~\ref{fig:refinement_example} and Fig.~\ref{fig:disentanglement}

\paragraph{Color-conditioning accuracy.}
Table~\ref{tab:color_fidelity} reports color fidelity using two complementary metrics. We primarily focus on Euclidean distance in the a–b chromaticity plane, which isolates hue and saturation differences while ignoring lighting, making it better aligned with our goal of precise parametric color control independent of illumination and shading. 
Under this metric, \modelname consistently outperforms all competing models for both $K = 5$ and $K = 8$, achieving the lowest mean, median, and 90th-percentile errors, and indicating both superior average accuracy and substantially fewer severe failures.
We also report CIEDE2000 ($\Delta E_{00}$), which penalizes lightness variation; some baselines achieve lower scores via more uniform lighting, whereas \modelname preserves accurate chromaticity under realistic lighting.

\section{Conclusion}

In this work, we introduced \modelname, a large-scale text-to-image model that enables precise control over object location, size, and color, through explicit numeric bounding boxes and RGB values.
\modelname directly addresses the parametric gap between descriptive language and the deterministic numeric control required in professional workflows, demonstrating that such precision can be achieved purely through large-scale training on enriched structured captions, without architectural modifications or inference-time optimization. More broadly, \modelname highlights the power of structured intermediate representations as a bridge between user intent and generative rendering. By translating natural-language prompts into a parametric schema that supports direct numeric manipulation, our framework enables intuitive interactive interfaces, such as object repositioning and precise color selection, while maintaining global scene coherence. This approach suggests a path toward programmable, professional-grade image synthesis systems that integrate additional precise attributes, moving beyond descriptive prompting toward truly controllable generative modeling.

\bibliographystyle{unsrtnat}
\bibliography{main}
\newpage 

\appendix
\twocolumn[{
\section{Additional Refinement Examples}
\vspace{0.5em}
\centering
\setlength{\tabcolsep}{4pt}
\begin{tabular}{ccc}
  \includegraphics[width=0.30\textwidth]{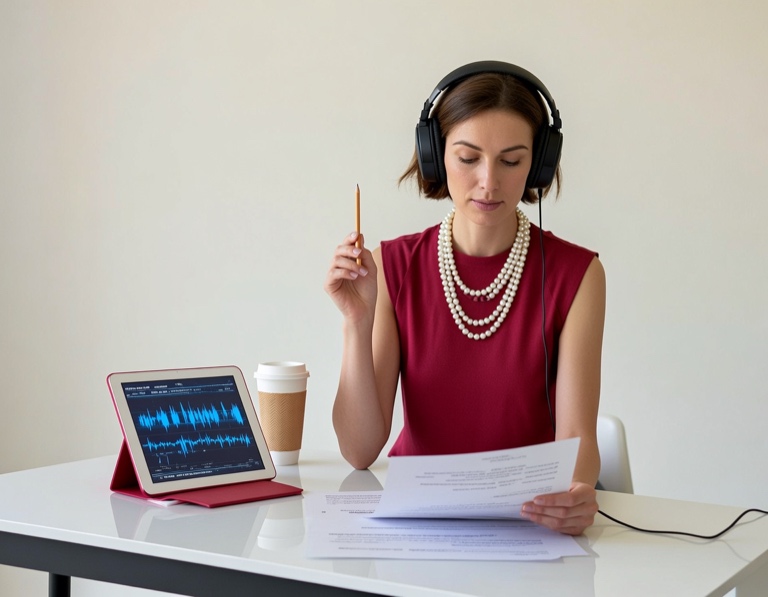} &
  \includegraphics[width=0.30\textwidth]{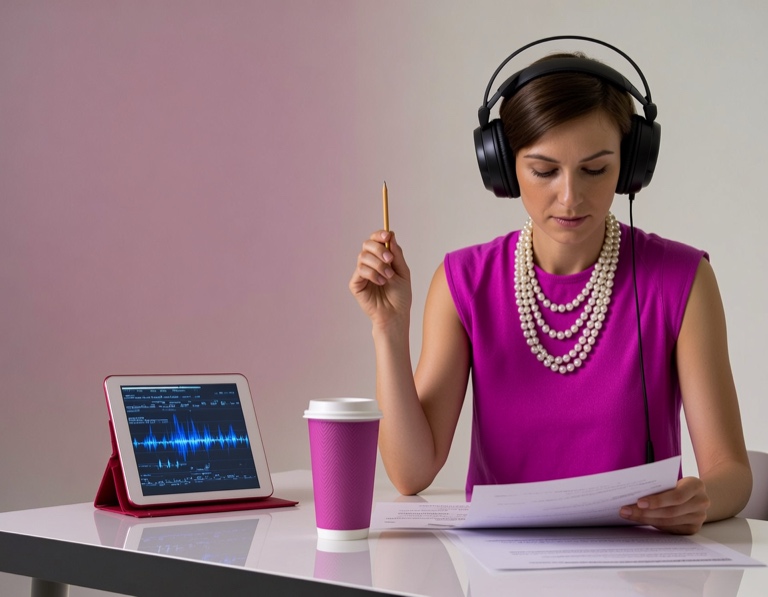} &
  \includegraphics[width=0.30\textwidth]{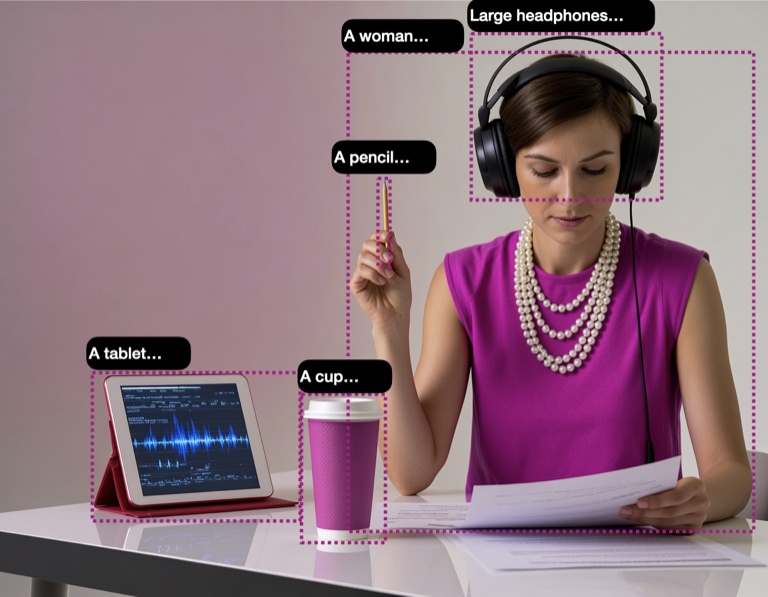} \\[3pt]
  \includegraphics[width=0.30\textwidth]{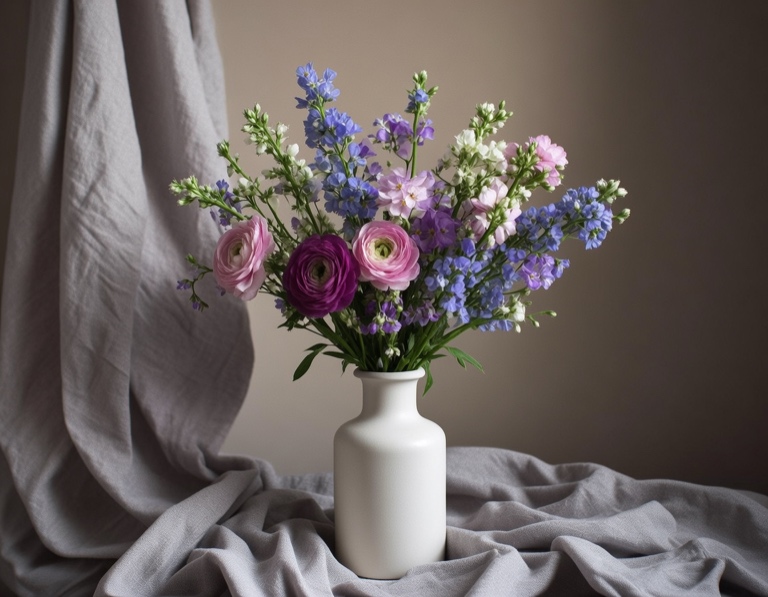} &
  \includegraphics[width=0.30\textwidth]{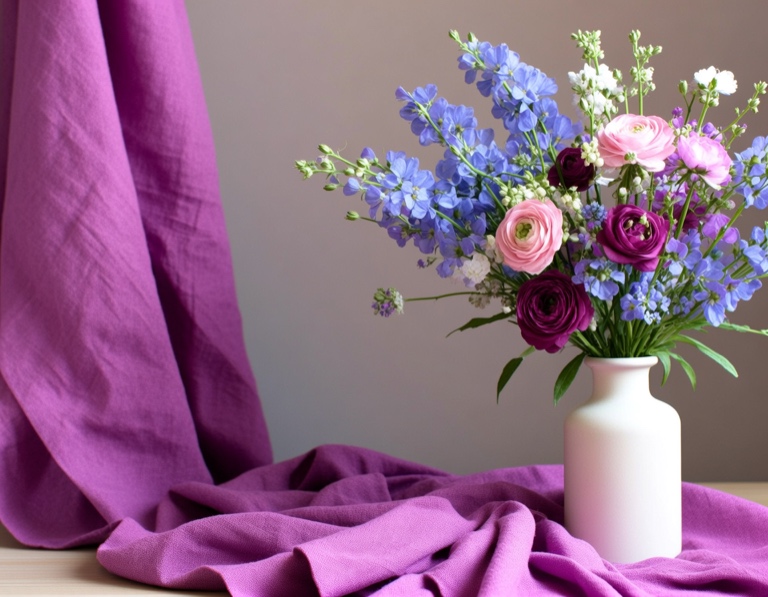} &
  \includegraphics[width=0.30\textwidth]{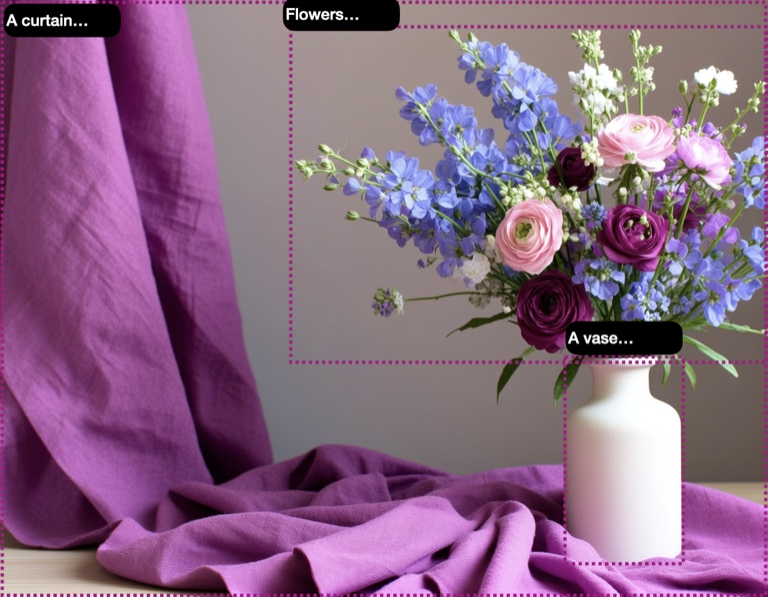} \\[3pt]
  \includegraphics[width=0.30\textwidth]{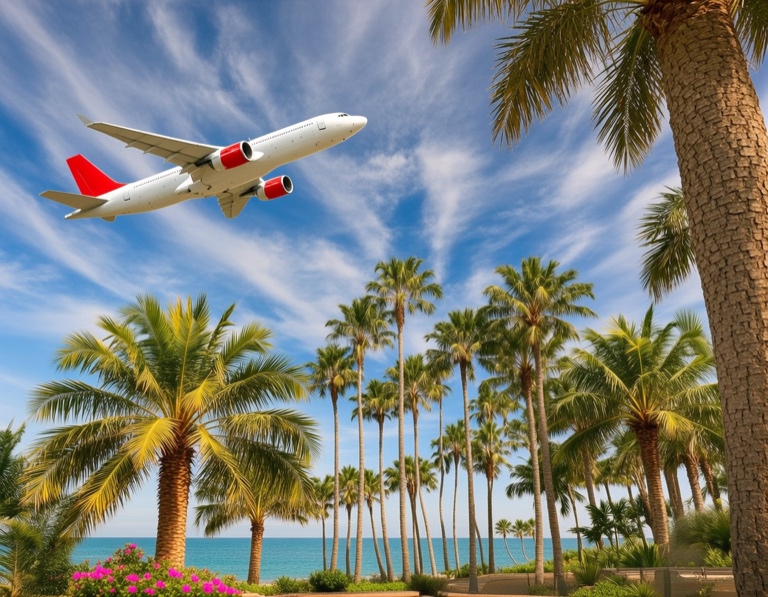} &
  \includegraphics[width=0.30\textwidth]{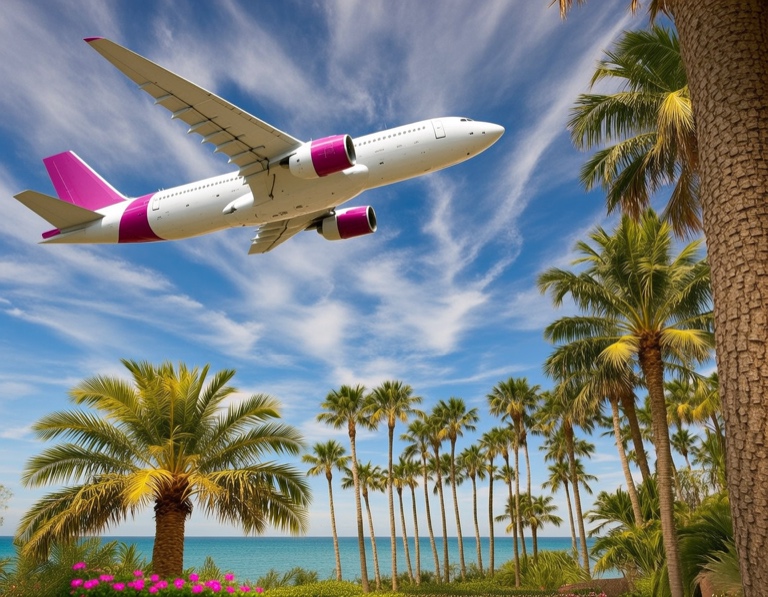} &
  \includegraphics[width=0.30\textwidth]{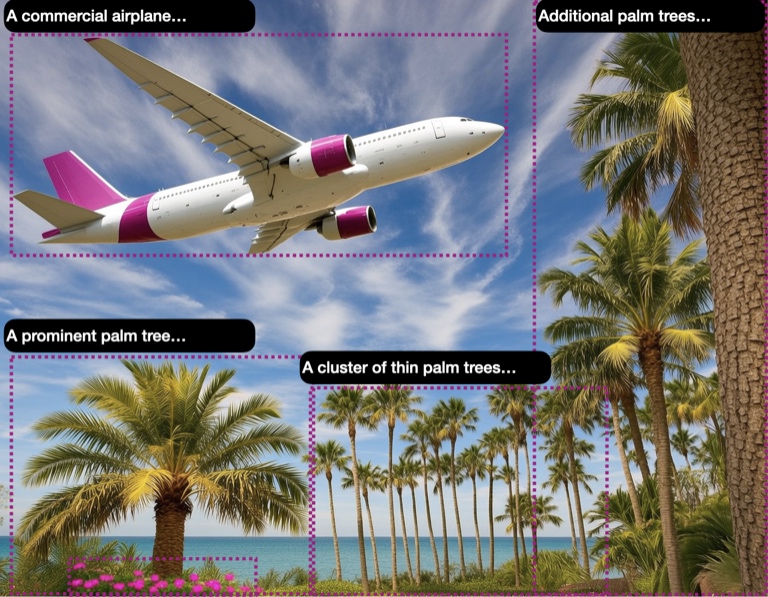} \\[3pt]
  {\bf Original} & {\bf Refined} & {\bf Refined with overlaid bounding boxes}
\end{tabular}

\vspace{4pt}
\captionof{figure}{\textbf{Refinement via structured parametric editing.}
The left column shows the original generations, while the middle column presents refined results obtained by editing the structured parametric caption and re-generating the image. In each example, both the numeric bounding boxes (object position and extent) and the object color are modified, explicitly enforcing the target color \textcolor[HTML]{DD20A7}{\#DD20A7}, resulting in updated spatial layout and appearance while preserving overall scene coherence. The right column overlays the exact numeric bounding boxes on the refined images, illustrating precise alignment with the edited parameters.}
\label{fig:refinement_example}
\vspace{0.5em}
}]

\end{document}